\documentclass[10pt,twocolumn,letterpaper]{article}

\usepackage{iccv}
\usepackage{times}
\usepackage{epsfig}
\usepackage{graphicx}
\usepackage{amsmath}
\usepackage{amssymb}

\usepackage[accsupp]{axessibility}  

\usepackage{booktabs}
\usepackage{hhline}

\usepackage{amssymb}
\usepackage{pifont}
\newcommand{\cmark}{\ding{51}}%
\newcommand{\xmark}{\ding{55}}%

\usepackage[pagebackref=true,breaklinks=true,letterpaper=true,colorlinks,bookmarks=false]{hyperref}

\iccvfinalcopy 

\def\iccvPaperID{9187} 

\ificcvfinal\fi

\begin{document}

\title{Pixel Difference Networks for Efficient Edge Detection}

\author{Zhuo Su$^{1,\ast}$\;\;\;\;\;
Wenzhe Liu$^{2,}$\thanks{Equal contributions.\;\;\;$\dagger$ Corresponding author: \href{http://lilyliliu.com}{http://lilyliliu.com}}\;\;\;\;\;
Zitong Yu$^1$\;\;\;\;\;
Dewen Hu$^2$\;\;\;\;\;
Qing Liao$^3$\;\;\;\;\;
Qi Tian$^4$\\
Matti Pietik{\"a}inen$^1$\;\;\;\;\;
Li Liu$^{2,1,\dagger}$\\
$^1$Center for Machine Vision and Signal Analysis, University of Oulu, Finland\\
$^2$National University of Defense Technology, China\\
$^3$Harbin Institute of Technology (Shenzhen), China\;\;\;\;\;$^4$Xidian University, China\\
{\tt\small \{zhuo.su, zitong.yu, matti.pietikainen, li.liu\}@oulu.fi}\\ 
{\tt\small \{liuwenzhe15, dwhu\}@nudt.edu.cn, liaoqing@hit.edu.cn, wywqtian@gmail.com}
}


\maketitle
\ificcvfinal\thispagestyle{empty}\fi

\begin{abstract}


Recently, deep Convolutional Neural Networks (CNNs) can achieve human-level performance in edge detection with the rich and abstract edge representation capacities. However, the high performance of CNN based edge detection is achieved with a large pretrained CNN backbone, which is memory and energy consuming. In addition, it is surprising that the previous wisdom from the traditional edge detectors, such as Canny, Sobel, and LBP are rarely investigated in 
the rapid-developing deep learning era. 
To address these issues, 
we propose a simple, lightweight yet effective architecture named Pixel Difference Network (PiDiNet) for efficient edge detection. 
PiDiNet adopts novel pixel difference convolutions that integrate the traditional edge detection operators into the popular convolutional operations in modern CNNs for enhanced performance on the task, which enjoys the best of both worlds.
Extensive experiments on BSDS500, NYUD, and Multicue are provided to demonstrate its effectiveness, and its high training and inference efficiency. Surprisingly, when training from scratch with only the BSDS500 and VOC datasets, PiDiNet can surpass the recorded result of human perception (0.807 vs. 0.803 in ODS F-measure) on the BSDS500 dataset with 100 FPS and less than 1M parameters. A faster version of PiDiNet with less than 0.1M parameters can still achieve comparable performance among state of the arts
with 200 FPS.
Results on the NYUD and Multicue datasets show similar observations. 
The codes are available at  \href{https://github.com/zhuoinoulu/pidinet}{https://github.com/zhuoinoulu/pidinet}.



\end{abstract}

\section{Introduction}
\label{sec:intro}

Edge detection has been a longstanding, fundamental low-level problem in computer vision~\cite{canny1986computational}. Edges and object boundaries play an important role in various higher-level computer vision tasks such as object recognition and detection \cite{liu2020deep,ferrari2007groups}, object proposal generation \cite{cheng2014bing,uijlings2013selective}, image editing \cite{elder1998imageediting}, and image segmentation \cite{muthukrishnan2011edgeimageseg,bertasius2016semantic}. Therefore, recently, the edge detection problem has also been revisited and injected new vitality due to the renaissance of deep learning~\cite{bertasius2015deepedge,kokkinos2015deepboundary,shen2015deepcontour,xie2017holistically,wang2017ced,liu2019richer} .

\begin{figure}[t!]
    \centering
    \includegraphics[width=0.98\linewidth]{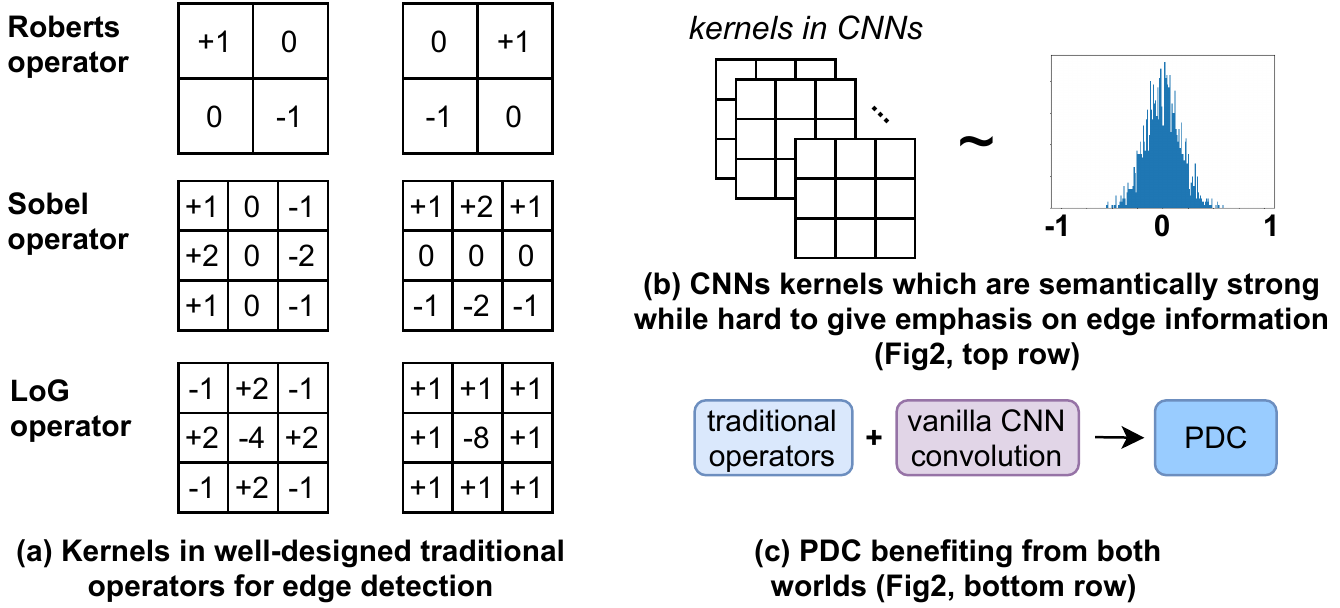}
    \caption{PDC benefits from both worlds with proper integration of traditional operators and modern CNNs.}
    \label{fig:figure1}
\end{figure}

\begin{figure*}[t!]
    \centering
    \includegraphics[width=0.98\linewidth]{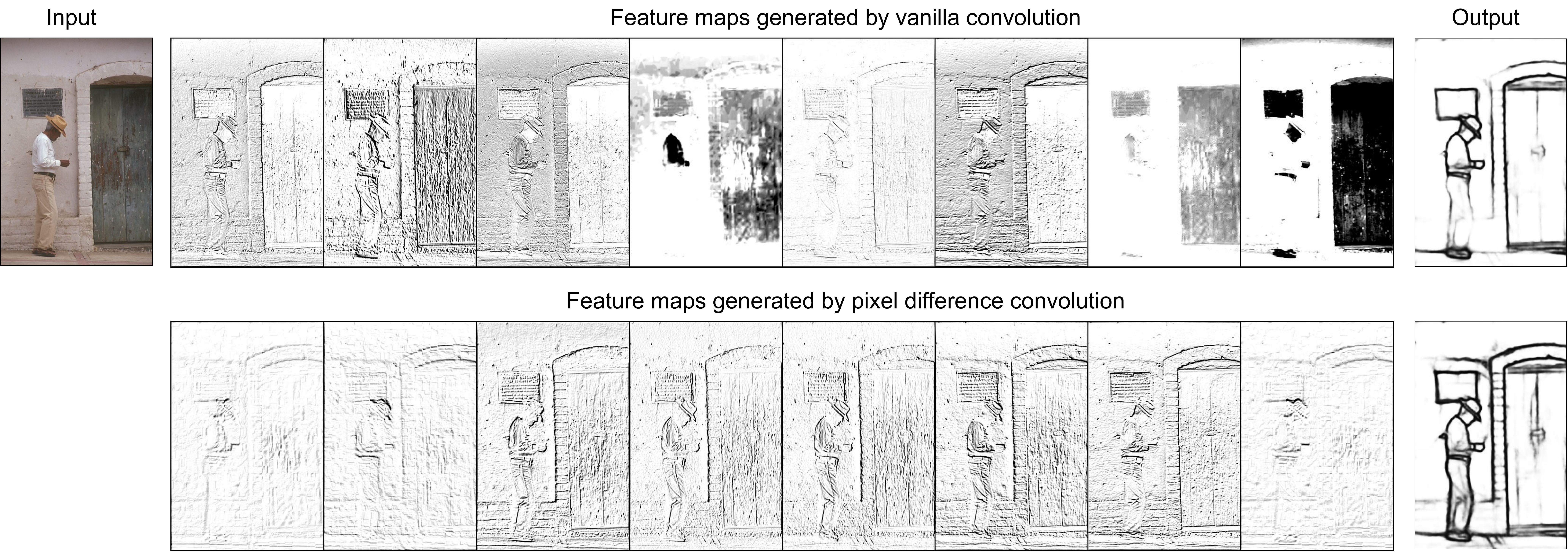}
    \caption{PiDiNet configured with pixel difference convolution (PDC) \emph{vs.} the baseline with vanilla convolution. Both models were trained only using the BSDS500 dataset. Compared with vanilla convolution, PDC can better capture gradient information from the image that facilitates edge detection.}
    \label{fig:figure2}
\end{figure*}

The main goal of edge detection is identifying sharp image brightness changes such as discontinuities in intensity, color, or texture~\cite{torre1986edge}. Traditionally, edge detectors based on image gradients or derivatives information are popular choices.
Early classical methods use the first or second order derivatives (\emph{e.g.}, Sobel~\cite{sobel19683x3}, Prewitt~\cite{prewitt1970object}, Laplacian of Gaussian (LoG), Canny~\cite{canny1986computational}, \emph{etc.}) for basic edge detection. Later learning based methods~\cite{hallman2015oef,dollar2014se} further utilize various gradient information~\cite{xiaofeng2012scg, martin2004pb,fowlkes2002learningpb,gupta2014seng} to 
produce more accurate boundaries.

Due to the capability of automatically learning rich representations of
data with hierarchical levels of abstraction, deep CNNs have brought tremendous progress for various computer vision tasks including edge detection and are still rapidly developing. Early deep learning based edge detection models
construct CNN architectures as classifiers to predict the edge probability of an input image patch~\cite{bertasius2015deepedge,shen2015deepcontour,bertasius2015hfl}.  
Building on top of fully convolutional networks~\cite{long2015fully}, HED~\cite{xie2017holistically} performs end-to-end edge detection by leveraging multilevel image features with rich hierarchical information guided by deep supervision, and achieves state-of-the-art performance. Other similar works include~\cite{yang2016cedn,kokkinos2015deepboundary,maninis2016cob,wang2017ced,xu2018amhnet,liu2019richer,deng2018lpcb,he2019bidirectional}.



However, integration of traditional edge detectors with modern CNNs were rarely investigated. The former were merely utilized as auxiliary tools to extract candidate edge points in some prior approaches~\cite{bertasius2015hfl,bertasius2015deepedge}.
Intuitively, edges manifest diverse specific patterns like straight lines, corners, and ``X'' junctions. On one hand, traditional edge operators like those shown in Fig. 1 are inspired by these intuitions, and based on gradient computing which encodes important gradient information for edge detection by explicitly calculating pixel differences. However, these handcrafted edge operators or learning based edge detection algorithms are usually not powerful enough due to their shallow structures. On the other hand, modern CNNs can learn rich and hierarchical image representations, where vanilla CNN kernels serve as probing local image patterns. Nevertheless, CNN kernels are optimized by starting from random initialization which has no explicit encoding for gradient information, making them hard to focus on edge related features.


\begin{table}[t!]
\caption{Comparison between ours and some leading edge detection models in terms of efficiency and accuracy. The multiply-accumulates (MACs) are calculated based on a 200$\times$200 image, FPS and ODS \emph{F-measure} are evaluated on the BSDS500 test set.}
\begin{center}
\setlength{\tabcolsep}{0.015\linewidth}
\resizebox*{\linewidth}{!}{
\begin{tabular}{lccccc}
\toprule[1pt]
 & \emph{HED}~\cite{xie2017holistically} & \emph{RCF}~\cite{liu2019richer} & \emph{BDCN}~\cite{he2019bidirectional} & \emph{PiDiNet} & \emph{PiDiNet(tiny)} \\
\hline
Params & 14.7M & 14.8M & 16.3M & 710K & 73K \\
MACs & 22.2G & 16.2G & 23.2G & 3.43G & 270M \\
Throughput & 78FPS & 67FPS & 47FPS & 92FPS & 215FPS\\
Pre-training & ImageNet & ImageNet & ImageNet & No & No \\
ODS \emph{F-measure} & 0.788 & 0.806 & 0.820 & 0.807 & 0.787 \\
\bottomrule[1pt]
\end{tabular}
}
\end{center}
\label{table:table1}
\vspace{-0.2in}
\end{table}

We believe a new type of convolutional operation can be derived, to satisfy the following needs. Firstly, it can easily capture the image gradient information that facilitates edge detection, and the CNN model can be more focused with the release of burden on dealing with much unrelated image features. Secondly, the powerful learning ability of deep CNNs can still be preserved, to extract semantically meaningful representations, which lead to robust and accurate edge detection. In this paper, we propose pixel difference convolution (PDC), where the pixel differences in the image are firstly computed, and then convolved with the kernel weights to generate output features (see Fig.~\ref{fig:pdc}). We show PDC can effectively improve the quality of the output edge maps, as illustrated
in Fig.~\ref{fig:figure2}. 

On the other hand, leading CNN based edge detectors suffer from the deficiencies as shown in Table~\ref{table:table1}: being memory consuming with big model size, being energy hungry with high computational cost, running inefficiency with low throughput and label inefficiency with the need of model pre-training on large scale dataset. 
This is due to the fact that the annotated data available for training edge detection models is limited, and thus a well pretrained (usually large) backbone is needed.
For example, the widely adopted routine is to use the large VGG16~\cite{simonyan2014very} architecture that was trained on the large scale ImageNet dataset~\cite{deng2009imagenet}. 


It is important to develop a lightweight structure, to achieve a better trade-off between accuracy and efficiency for edge detection. With pixel difference convolution, inspired by~\cite{he2016residual,howard2017mobilenets}, we build a new end-to-end architecture, namely Pixel Difference Network (PiDiNet) to solve the mentioned issues in one time. Specifically, PiDiNet consists of an efficient backbone and an efficient task-specific side structure (see Fig.~\ref{fig:arch}), able to do robust and accurate edge detection with high efficiency.


\section{Related Work}


\noindent \textbf{Using Traditional Edge Detectors to Help Deep CNN Models for Edge Detection.} \quad Canny~\cite{canny1986computational} and SE~\cite{dollar2014se} edge detectors are usually used to extract candidate contour points before applying the CNN model for contour/non-contour prediction~\cite{bertasius2015deepedge, bertasius2015hfl}. The candidate points can be also used as auxiliary relaxed labels for better training the CNN model~\cite{liu2016relaxed}. Instead of relying on the edge information from the hand-crafted detectors, PDC directly integrates the gradient information extraction process into the convolutional operation, which is more compact and learnable.

\vspace{0.3em}
\noindent \textbf{Lightweight Architectures for Edge Detection.} \quad Recently, efforts have been made to design lightweight architectures for efficient edge detection~\cite{wibisono2020fined,wibisono2020traditional,poma2020dense}. Some of them may not need a pretrained network based on large scale dataset~\cite{poma2020dense}. Although being compact and fast, the detection accuracies with these networks are unsatisfactory. Alternatively, lightweight architectures for other dense prediction tasks~\cite{gao2020100k,wu2020cgnet,paszke2016enet,li2019dabnet,mehta2019espnetv2,yu2018bisenet} and multi-task learning~\cite{kokkinos2017ubernet, liu2020dynamicintegration} may also benefit edge detection. However, the introduced sophisticated multi-branch based structures may lead to running inefficiency. Instead, we build a backbone structure which only uses a simple shortcut~\cite{he2016residual} as the second branch for the convolutional blocks.

\vspace{0.3em}
\noindent \textbf{Integrating Traditional Operators.}  \quad The proposed PDC is mostly related to the recent central difference convolution (CDC)~\cite{yu2020cdc,yu2020fas,yu2021dual,yu2021searching} and local binary convolution (LBC)~\cite{juefei2017lbc}, of which both derive from local binary patterns (LBP)~\cite{ojala2002lbp} and involve calculating pixel differences during convolution. LBC uses a set of predefined sparse binary filters to generalize the traditional LBP, focusing on reducing the network complexity. CDC further proposes to use learnable weights to capture image gradient information for robust face anti-spoofing. CDC can be seen as one instantiated case of the proposed PDC (\emph{i.e.}, Central PDC), where the central direction is considered, as we will introduce in Section~\ref{sec:pdc}. Like CDC, PDC uses learnable filters while being more general and flexible to capture rich gradient information for edge detection. On the other hand, Gabor convolution~\cite{luan2018gabor} encodes the orientation and scale information in the convolution kernels by multiplying the kernels with a group of Gabor filters, while PDC is more compact without any auxiliary traditional feature filters.



\section{Pixel Difference Convolution}
\label{sec:pdc}

\begin{figure}[t!]
    \centering
    \includegraphics[width=1\linewidth]{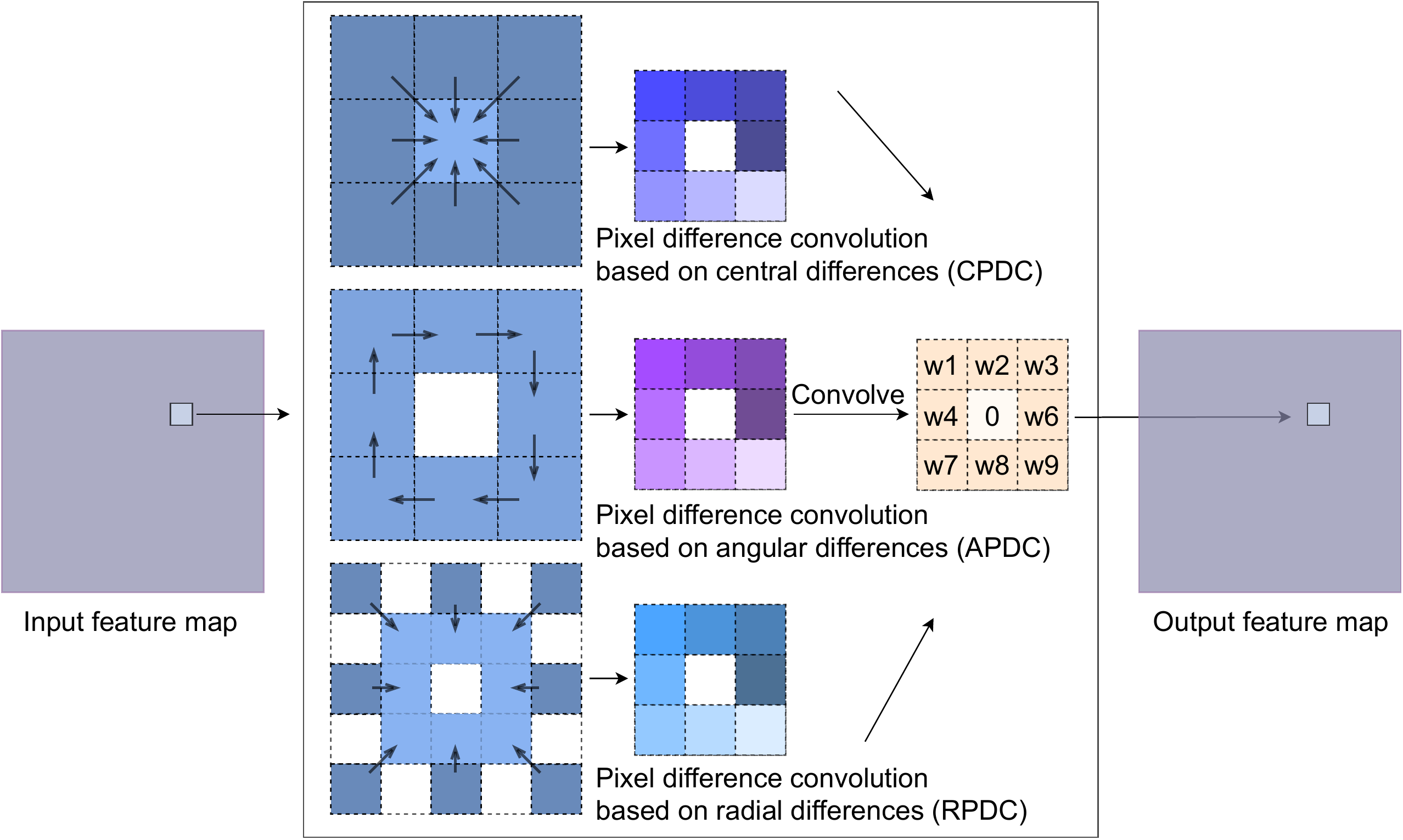}
    \caption{Three instances of pixel difference convolution derived from extended LBP descriptors~\cite{liu2011sorted, liu2012extended, su2019bird}. One can derive other instances by designing the picking strategy of the pixel pairs.}
    \label{fig:pdc}
\end{figure}

The process of pixel difference convolution (PDC) is pretty similar to that of vanilla convolution,
where the original pixels in the local feature map patch covered by the convolution kernels are replaced by pixel differences, when conducting the convolutional operation. The formulations of vanilla convolution and PDC can be written as:
\begin{align}
    y &= f(\pmb{x}, \pmb{\theta}) = \sum_{i=1}^{k\times k}w_{i}\cdot x_{i}, \;\;\;\;\;\;\; \text{(vanilla convolution)} \\
    y &= f(\triangledown\pmb{x}, \pmb{\theta}) = \sum_{(x_i, x_i')\in \pmb{\mathcal{P}}}w_{i}\cdot (x_i - x_i'), \;\;\;\;\;\;\, \text{(PDC)} \label{eq: pdc}
\end{align}
where, $x_i$ and $x_i'$ are the input pixels, $w_i$ is the weight in the $k \times k$ convolution kernel. $\pmb{\mathcal{P}} = \{(x_1, x_1'), (x_2, x_2'), ..., (x_m, x_m')\}$ is the set of pixel pairs picked from the current local patch, and $m\le k\times k$. 

To capture rich gradient information, the pixel pairs can be selected according to different strategies, which can be inspired from the numerous traditional feature descriptors. Here, we utilize the ideas from the work in~\cite{ojala2002lbp,liu2012extended,su2019bird}, where the local binary pattern (LBP) and its robust variants, extended LBP (ELBP), were used to encode pixel relations from varying directions (angular and radial). Specifically, ELBP are obtained by firstly calculating the pixel differences within a local patch (from $m$ pixel pairs), resulting in a pixel difference vector, and then binarizing the vector to create an $m$-length 0/1 code. Then, the bag-of-words technique~\cite{liu2019bow} is usually used to calculate the code distribution (or histogram), which is regarded as the image representation. In ELBP, the angular and radial directions were will demonstrated to help encode potential discriminative image cues and be complementary for increasing the feature representational capacity for various computer vision tasks, such as texture classification~\cite{liu2012extended,liu2011sorted} and face recognition~\cite{su2019bird}.

By integrating ELBP with CNN convolution, we derive three types of PDC instances as shown in Fig.~\ref{fig:pdc}, in which we name them as central PDC (CPDC), angular PDC (APDC) and radial PDC (RPDC) respectively. The pixel pairs in the local patch is easy to understand. For example, for the APDC with kernel size $3\times 3$, we create 8 pairs in the angular direction in the $3\times 3$ local patch (thus $m=8$), then the pixel differences obtained from the pairs are convolved with the kernel by doing an element-wise multiplication with the kernel weights, followed by a summation, to generate the value in the output feature map.

The derived PDC instances based on ELBP can be seen as an extension of ELBP that are more flexible and learnable. Although being powerful, the original ELBP codes are discrete with limited representative ability. While the useful encodings of pixel relations in PDC will be preserved in the trained convolution kernels, as during the training process of CNN, the convolution kernels will be encouraged to have higher 
inner product
with those important encodings, in order to create higher activation responses\footnote{Usually, higher activation responses are considered to be more salient, as adopted in many network pruning methods~\cite{han2015deepcompression,su2020dynamic}}. By training from abundant of data, PDC is able to automatically learn rich representative encodings for the task. 

\begin{figure}[h]
    \centering
    \includegraphics[width=0.8\linewidth]{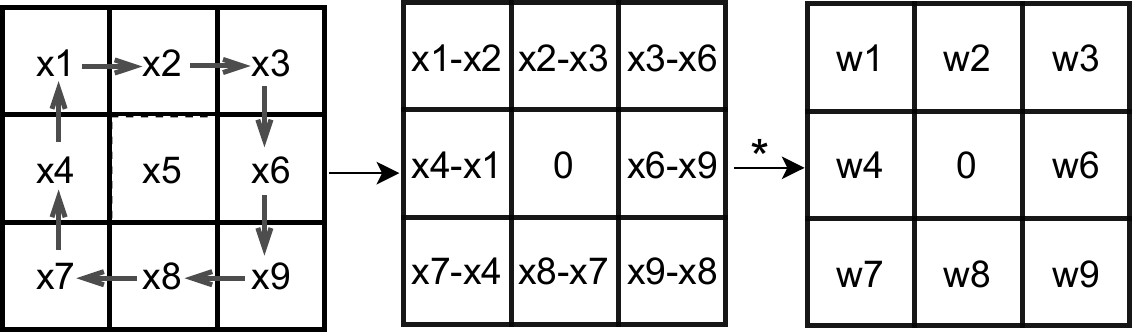}
    \caption{Selection of pixel pairs and convolution in APDC.}
    \label{fig:apdc}
\end{figure}

\vspace{0.3em}
\noindent \textbf{Converting PDC to Vanilla Convolution.} \quad
According to Eq.~\ref{eq: pdc}, one may notice that the computational cost and memory footprint by PDC are doubled compared with the vanilla counterpart. However, once the convolution kernels have been learnt, PDC layers can be converted to vanilla convolutional layers by instead saving the differences of the kernel weights in the model, according to the locations of the selected pixel pairs. In this way, the efficiency is maintained during inference. Taking APDC as an example (Fig.~\ref{fig:apdc}), conversion is done with the following equations:
{\small 
\begin{align}
    y &= w_{1}\cdot (x_1 - x_2) + w_2\cdot (x_2 - x_3)+w_3\cdot (x_3 - x_6) + ... \nonumber\\
    &=(w_1 - w_4)\cdot x_1 + (w_2 - w_1)\cdot x_2 + (w_3 - w_2)\cdot x_3 + ...\nonumber\\
    &=\hat{w}_1\cdot x_1 + \hat{w}_2\cdot x_2 + \hat{w}_3\cdot x_3 + ... =\sum \hat{w}_i\cdot x_i.
\end{align}
}

It is worth mentioning that we can also use this tweak to speed up the training process, where the differences of kernel weights are firstly calculated, followed by the convolution with the untouched input feature maps.
We have illustrated more details in the appendix.

\section{PiDiNet Architecture}

\begin{figure*}[t!]
    \centering
    \includegraphics[width=0.98\linewidth]{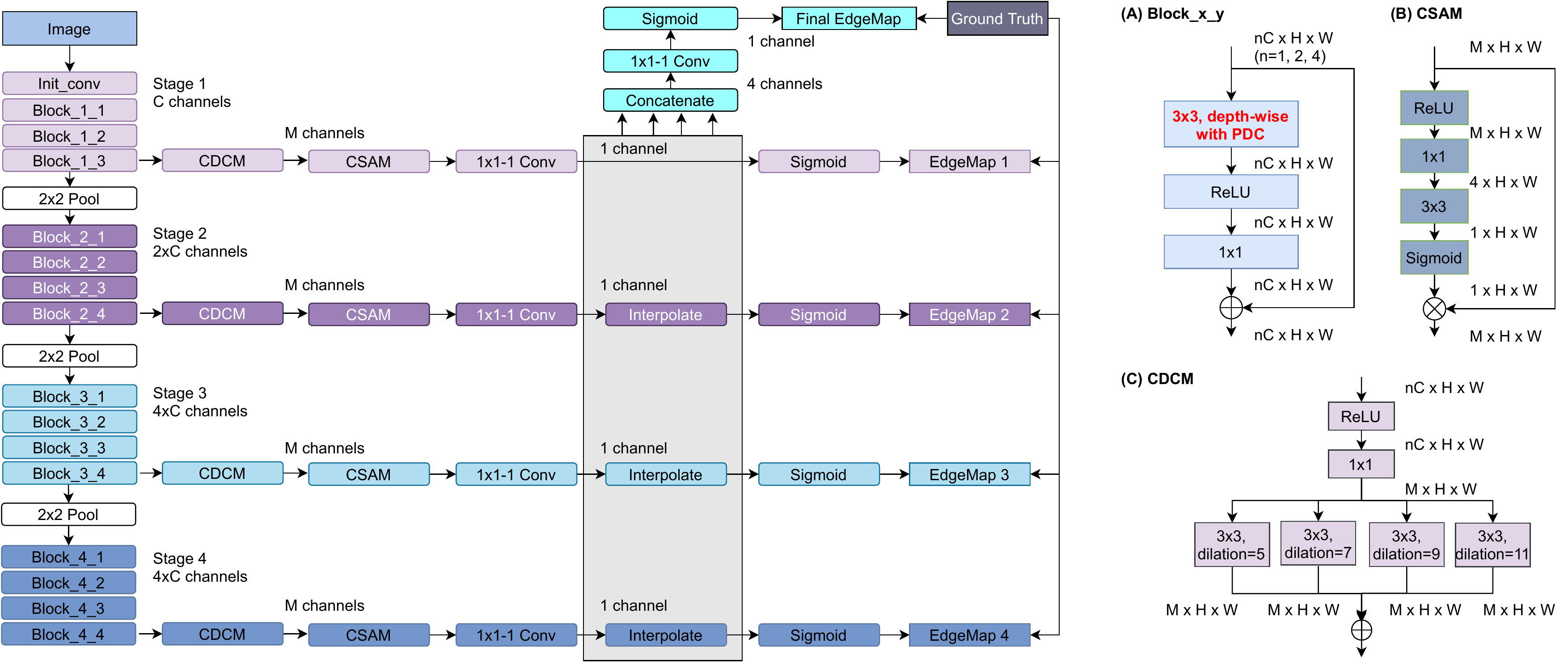}
    \caption{PiDiNet architecture.}
    \label{fig:arch}
\end{figure*}

As tried by some prior works~\cite{wibisono2020fined,poma2020dense,wibisono2020traditional}, we believe it is both necessary and feasible to solve the inefficiency issues mentioned in Section~\ref{sec:intro} in one time by building an architecture with small model size and high running efficiency, and can be trained from scratch using limited datasets for effective edge detection. We construct our architecture with the following parts (Fig.~\ref{fig:arch}).

\vspace{0.3em}
\noindent \textbf{Efficient Backbone.} \quad The building principle for the backbone is to make the structure slim while own high running efficiency. Thus we do not consider the sophisticated multi-branch lightweight structures proposed for many other tasks~\cite{gao2020100k,mehta2019espnetv2,yu2018bisenet}, since they may not appeal to parallel implementation~\cite{ma2018shufflenetv2}, leading to unsatisfactory efficiency for the edge detection task. Inspired from \cite{he2016residual} and \cite{howard2017mobilenets}, we use the separable depth-wise convolutional structure with a shortcut for fast inference and easy training. The whole backbone has 4 stages
and max pooling layers are among them for down sampling.
Each stage has 4 residual blocks (except the first stage that has an initial convolutional layer and 3 residual blocks). The residual path in each block includes a depth-wise convolutional layer, a ReLU layer, and a point-wise convolutional layer sequentially. 
The number of channels in each stage is reasonably small to avoid big model size ($C$, $2\times C$, $4\times C$ and $4\times C$ channels for stage 1, 2, 3, and 4 respectively).

\vspace{0.3em}
\noindent \textbf{Efficient Side Structure.} \quad 
To learn rich hierarchical edge representation, we also use the side structure as in~\cite{xie2017holistically} to generate an edge map from each stage respectively, based on which a side loss is computed with the ground truth map to provide deep supervision~\cite{xie2017holistically}. 
To refine the feature maps, beginning from the end of each stage, we firstly build a compact dilation convolution based module (CDCM) to enrich multi-scale edge information, which takes the input with $n\times C$ channels, and produces $M$ ($M < C$) channels in the output to relieve the computation overhead, followed by a compact spatial attention module (CSAM) to eliminate the background noise. After that, a $1\times 1$ convolutional layer further reduces the feature volume to a single channel map, which is then interpolated to the original size followed by a Sigmoid function to create the edge map. The final edge map, which is used for testing, is created by fusing the 4 single channel feature maps with a concatenation, a convolutional layer and a Sigmoid function. 

The detailed structure information can be seen in Fig.~\ref{fig:arch}, noting that we do not use any normalization layers for simplicity since the resolutions of the training images are not uniform. The obtained architecture is our baseline. By replacing the vanilla convolution in the 
$3\times 3$ depth-wise convolutional layer in the residual blocks with PDC, we get the proposed PiDiNet. 

\vspace{0.3em}
\noindent \textbf{Loss Function.} \quad We adopt the annotator-robust loss function proposed in~\cite{liu2019richer} for each generated edge map (including the final edge map). For the $i$th pixel in the $j$th edge map with value $p_i^j$, the loss is calculated as:
{\small \begin{equation}
    l_i^j = \begin{cases} 
    \alpha\cdot \log (1 - p_i^j) &\text{if } y_i  = 0 \\
    0 &\text{if } 0 < y_i < \eta \\
    \beta\cdot \log p_i^j &\text{otherwise},
    \end{cases}
\end{equation}}
where $y_i$ is the ground truth edge probability, $\eta$ is a pre-defined threshold, meaning that a pixel is discarded and not considered to be a sample when calculating the loss if it is marked as positive by fewer than $\eta$ of annotators to avoid confusing, $\beta$ is the percentage of negative pixel samples and $\alpha = \lambda\cdot (1-\beta)$. After all, the total loss is $L = \sum_{i,j}l_i^j$.

\section{Experiments}
\label{sec:experiments}

\subsection{Datasets and Implementation}
\noindent \textbf{Experimental Datasets.} \quad We evaluate the proposed PiDiNet on three widely used datasets, namely, BSDS500~\cite{arbelaez2010bsds}, NYUD~\cite{shi2000nyud}, and Multicue~\cite{mely2016multicue}. The experimental settings about data augmentation and configuration on the three datasets follow~\cite{xie2017holistically,liu2019richer,he2019bidirectional} and the details are given below. BSDS500 consists of 200, 100, and 200 images in the training set, validation set, and test set respectively. Each image has 4 to 9 annotators. 
Training images in the dataset are augmented with flipping (2$\times$), scaling (3$\times$), and rotation (16$\times$), leading to a training set that is 96$\times$ larger than the unaugmented version. 
Like prior works~\cite{xie2017holistically,liu2019richer,he2019bidirectional},
the PASCAL VOC Context dataset~\cite{mottaghi2014voc}, which has 10K labeled images (and augmented to 20K with flipping), is also optionally considered in training. NYUD has 1449 pairs of aligned RGB and depth images which are densely labeled. There are 381, 414 and 654 images for training, validation, and test respectively. We combine the training and validation set and augment them with flipping (2$\times$), scaling (3$\times$), and rotation (4$\times$) to produce the training data. Multicue is composed of 100 challenging natural scenes and each scene contains a left- and right-view color sequences captured by a binocular stereo camera. The last frame of left-view sequences for each scene, which is labeled with edges and boundaries, is used in our experiments. We randomly split them to 80 and 20 images for training and evaluation respectively. The process is independently repeated twice more. The metrics are then recorded from the three runs. We also augment each training image with flipping (2$\times$), scaling (3$\times$), and rotation ($16\times$), then randomly crop them with size 500$\times$500.

\vspace{0.3em}
\noindent \textbf{Performance Metrics.} \quad During evaluation, \emph{F-measure} at both Optimal Dataset Scale (ODS) and Optimal Image Scale (OIS) are recorded for all datasets. Since efficiency is one of the main focuses in this paper, all the models are compared based on the evaluations from single scale images if not specified.

\vspace{0.3em}
\noindent \textbf{Implementation Details.} \quad Our implementation is based on the Pytorch library~\cite{paszke2019pytorch}. In detail, PiDiNet (and the baseline) is randomly initialized and trained for 14 epochs with Adam optimizer~\cite{kingma2014adam} with an initial learning rate 0.005, which is decayed in a multi-step way (at epoch 8 and 12 with decaying rate 0.1). If VOC dataset is used in training for evaluating BSDS500, we train 20 epochs and decay the learning rate at epoch 10 and 16. 
$\lambda$ is set to 1.1 for both BSDS500 and Multicue, and 1.3 for NYUD. The threshold $\eta$ is set to 0.3 for both BSDS500 and Multicue. 
No $\eta$ is needed for NYUD
since the images are singly annotated. 

\begin{table}[t!]
\caption{Possible configurations of PiDiNet. `C', `A', `R' and `V' indicate CPDC, APDC, RPDC and vanilla convolution respectively. `$\times$n' means repeating the pattern for $n$ times sequentially. For example, the baseline architectrue can be presented as ``[V]$\times$16'', and `C-[V]$\times$15' means using CPDC in the first block and vanilla convolutions in the later blocks. All the models are trained using BSDS500 training set and the VOC dataset, then evaluated on BSDS500 validation set.}
\begin{center}
\setlength{\tabcolsep}{0.05\linewidth}
\resizebox*{\linewidth}{!}{
\begin{tabular}{l||l|l|l}
\toprule[1pt]
Architecture & C-[V]$\times$15 & A-[V]$\times$15 & R-[V]$\times$15 \\
\hline
ODS / OIS & 0.775 / 0.794 & 0.774 / 0.794 & 0.774 / 0.792 \\
\hline
Architecture & [CVVV]$\times$4 & [AVVV]$\times$4 & [RVVV]$\times$4 \\
\hline
ODS / OIS & 0.773 / 0.792 & 0.771 / 0.790 & 0.772 / 0.791 \\
\hline
Architecture & [CCCV]$\times$4 & [AAAV]$\times$4 & [RRRV]$\times$4 \\
\hline
ODS / OIS & 0.772 / 0.791 & 0.775 / 0.793 & 0.771 / 0.787 \\
\hline
Architecture & [C]$\times$16 & [A]$\times$16 & [R]$\times$16 \\
\hline
ODS / OIS & 0.767 / 0.786 & 0.768 / 0.786 & 0.758 / 0.777 \\
\hline
Architecture & Baseline & \multicolumn{2}{l}{\textbf{[CARV]$\times$4 (PiDiNet)}} \\
\hline
ODS / OIS & 0.772 / 0.792 & \multicolumn{2}{l}{\textbf{0.776 / 0.795}} \\
\bottomrule[1pt]
\end{tabular}
}
\end{center}
\label{table:configuration}
\end{table}

\begin{table}[t!]
\caption{More comparisons between PiDiNet and the baseline architecture in multiple network scales by changing the nubmer of channels $C$ (see Fig.~\ref{fig:arch}). The models are trained using the BSDS500 training set, and evaluated on BSDS500 validation set.}
\begin{center}
\setlength{\tabcolsep}{0.05\linewidth}
\resizebox*{\linewidth}{!}{
\begin{tabular}{l|c|c}
\toprule[1pt]
Scale & Baseline (ODS / OIS) & PiDiNet (ODS / OIS) \\
\hline
Tiny (C=20) & 0.735 / 0.752 & \textbf{0.747 / 0.764} \\
\hline
Small (C=30) & 0.738 / 0.759 & \textbf{0.752 / 0.769} \\
\hline
Normal (C=60) & 0.736 / 0.751 & \textbf{0.757 / 0.776} \\
\bottomrule[1pt]
\end{tabular}
}
\end{center}
\label{table:morecomparison}
\end{table}

\begin{table}[t!]
\caption{Ablation on CDCM, CSAM and shortcuts. The models are trained with BSDS500 training set and VOC dataset, and evaluated on BSDS500 validation set.}
\begin{center}
\setlength{\tabcolsep}{0.08\linewidth}
\resizebox*{\linewidth}{!}{
\begin{tabular}{ccc|c}
\toprule[1pt]
CSAM & CDCM & Shortcuts & ODS / OIS \\
\hline
\xmark & \xmark & \cmark & 0.770 / 0.790 \\
\hline
\xmark & \cmark & \cmark & 0.775 / 0.793 \\
\hline
\cmark & \cmark & \cmark & \textbf{0.776 / 0.795} \\
\hline
\cmark & \cmark & \xmark & 0.734 / 0.755 \\
\bottomrule[1pt]
\end{tabular}
}
\end{center}
\label{table:moreablation}
\end{table}

\subsection{Ablation Study}
To demonstrate the effectiveness of PDC and to find the possibly optimal architecture configuration, we conduct our ablation study on the BSDS500 dataset, where we use the data augmented from the 200 images in the training set (optionally mixed with the VOC dataset) for training and record the metrics on the validation set. 

\vspace{0.3em}
\noindent \textbf{Architecture Configuration.} \quad We can replace the vanilla convolution with PDC in any block (we also regard the initial convolutional layer as a block in the context) in the backbone. Since there are 16 blocks, and a brute force search for the architecture configurations is not feasible, hence we only sample some of them as shown in Table~\ref{table:configuration} by gradually increasing the number of PDCs. We found replacing the vanilla convolution with PDC only in a single block can even have obvious improvement. More replacements with the same type of PDC may no longer give extra performance gain and instead degenerate the model. We conjecture that the PDC in the first block already obtains much gradient information from the raw image, and an abuse of PDC may even cause the model fail to preserve useful information. The extreme case is that when all the blocks are configured with PDC, the performance becomes worse than that of the baseline. The best configuration is `[CARV]$\times$4', which means combing the 4 types of convolutions sequentially in each stage, as different types of PDC capture the gradient information in different encoding directions. We will use this configuration in the following experiments. 

To further demonstrate the superiority of PiDiNet over the baseline, which only uses the vanilla convolution, we give more comparisons as shown in Table~\ref{table:morecomparison}. It constantly proves that PDC configured architectures outperform the corresponding vanilla convolution configured architectures.

\vspace{0.3em}
\noindent \textbf{CSAM, CDCM and Shortcuts.} \quad The effectiveness of CSAM, CDCM and residual structures are demonstrated in Table~\ref{table:moreablation}. The addition of shortcuts is simple yet important, as they can help preserve the gradient information captured by the previous layers. On the other hand, the attention mechanism in CSAM and dilation convolution in CDCM can give extra performance gains, while may also bring some computational cost. Therefore, they can be used to tradeoff between accuracy and efficiency. In the following experiments, we note PiDiNet without CSAM and CDCM as PiDiNet-L (meaning a more lightweight version).

\begin{figure}[t!]
    \centering
    \includegraphics[width=1\linewidth]{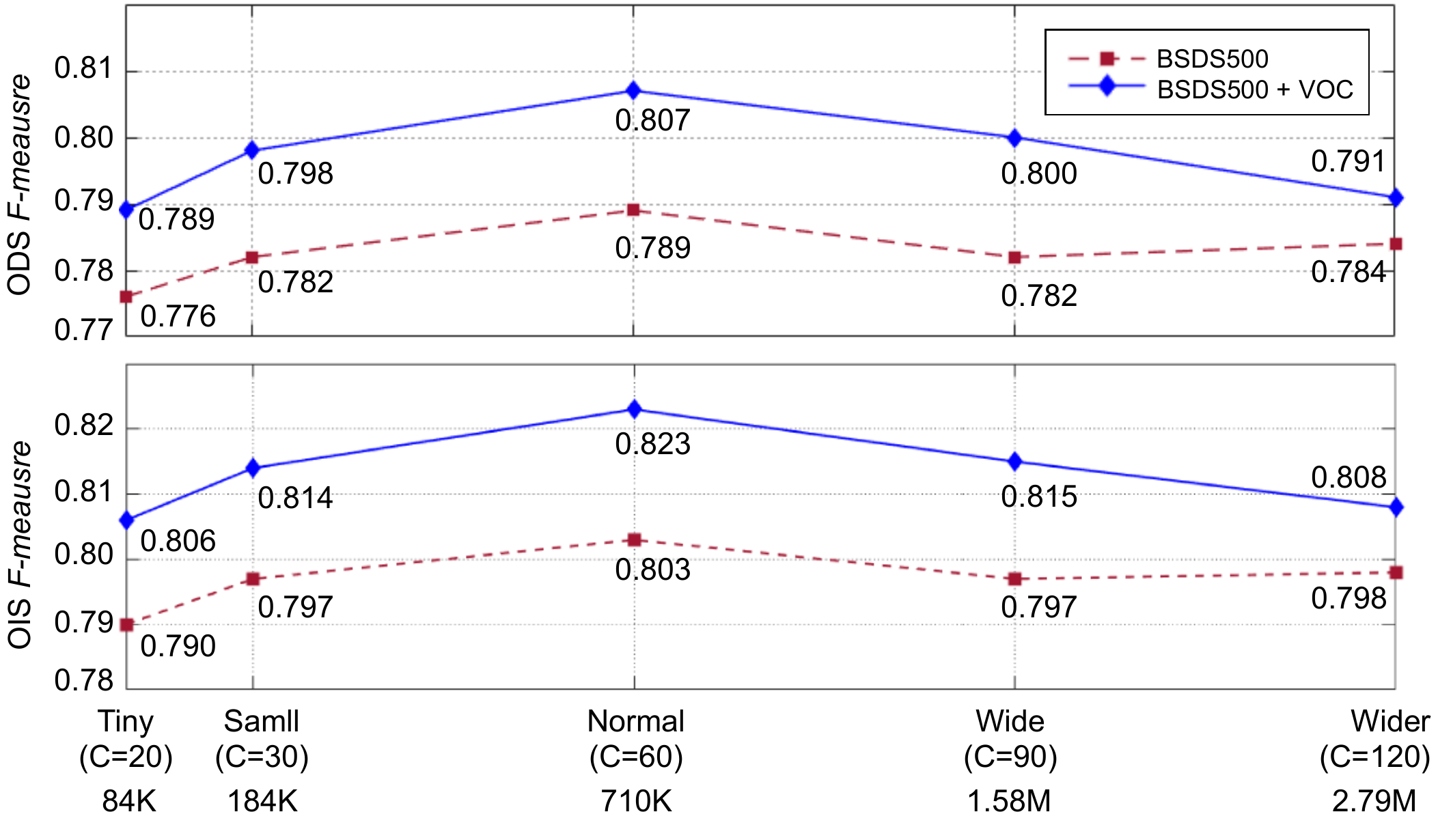}
    \caption{Exploration on the scalability of PiDiNet. The structure sizes are changed by slimming or widening the basic PiDiNet. Bottom row shows the number of parameters for each model. The models are trained with or without VOC dataset.}
    \label{fig:scalability}
\end{figure}

\begin{figure*}[t!]
    \centering
    \includegraphics[width=1\linewidth]{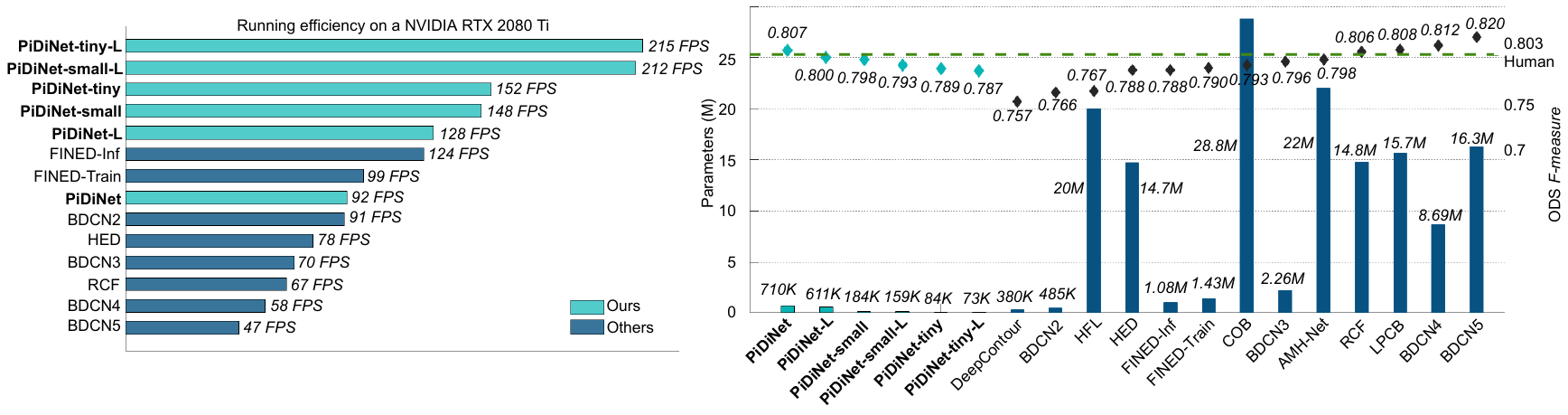}
    \caption{Comparison with other methods in terms of network complexity, running efficiency and detection performance (on BSDS500 dataset). The running speeds of FINED~\cite{wibisono2020fined} are cited from the original paper, and the rest are evaluated by our implementations}
    \label{fig:efficient}
\end{figure*}

\begin{table}[t!]
\caption{Comparison with other methods on BSDS500 dataset. $^\ddagger$ indicates the speeds with our implementations based on a NVIDIA RTX 2080 Ti GPU. $^\dagger$ indicates the cited GPU speeds.}
\begin{center}
\setlength{\tabcolsep}{0.04\linewidth}
\resizebox*{0.82\linewidth}{!}{
\begin{tabular}{l|c|c|c}
\toprule[1pt]
Method & ODS & OIS & FPS \\
\hline
Human & .803 & .803 & \\
\hline
Canny~\cite{canny1986computational} & .611 & .676 & 28 \\
Pb~\cite{martin2004pb} & .672 & .695 & - \\
SCG~\cite{xiaofeng2012scg} & .739 & .758 & - \\
SE~\cite{dollar2014se} & .743 & .763 & 12.5 \\
OEF~\cite{hallman2015oef} & .746 & .770 & 2/3 \\
\hline
DeepEdge~\cite{bertasius2015deepedge} & .753 & .772 & 1/1000$^\dagger$ \\
DeepContour~\cite{shen2015deepcontour} & .757 & .776 & 1/30$^\dagger$ \\
HFL~\cite{bertasius2015hfl} & .767 & .788 & 5/6$^\dagger$ \\
CEDN~\cite{yang2016cedn} & .788 & .804 & 10$^\dagger$ \\
HED~\cite{xie2017holistically} & .788 & .808 & 78$^\ddagger$ \\
DeepBoundary~\cite{kokkinos2015deepboundary} & .789 & .811 & -\\
COB~\cite{maninis2016cob} & .793 & .820 & - \\
CED~\cite{wang2017ced} & .794 & .811 & - \\
AMH-Net~\cite{xu2018amhnet} & .798 & .829 & - \\
RCF~\cite{liu2019richer} & .806 & .823 & 67$^\ddagger$ \\
LPCB~\cite{deng2018lpcb} & .808 & .824 & 30$^\dagger$ \\
BDCN~\cite{he2019bidirectional} & .820 & .838 & 47$^\ddagger$ \\
\hline
FINED-Inf~\cite{wibisono2020fined} & .788 & .804 & 124$^\dagger$ \\
FINED-Train~\cite{wibisono2020fined} & .790 & .808 & 99$^\dagger$ \\
\hline
Baseline & .798 & .816 & 96$^{\ddagger}$ \\
PiDiNet & .807 & .823 & 92$^{\ddagger,\ast}$ \\
PiDiNet-L & .800 & .815 & 128$^\ddagger$ \\
PiDiNet-Small & .798 & .814 & 148$^\ddagger$ \\
PiDiNet-Small-L & .793 & .809 & 212$^\ddagger$ \\
PiDiNet-Tiny & .789 & .806 & 152$^\ddagger$ \\
PiDiNet-Tiny-L & .787 & .804 & 215$^\ddagger$ \\
\bottomrule[1pt]
\multicolumn{4}{l}{\footnotesize{$^\ast$ ``PiDiNet'' is slightly slower than ``Baseline'' because RPDC is}}\\
\multicolumn{4}{l}{\footnotesize{a 5x5 convolution after conversion.}}
\end{tabular}
}
\end{center}
\label{table:bsds}
\vspace{-0.2in}
\end{table}

\subsection{Network Scalability}
\label{sec:scalability}
PiDiNet is highly compact with only 710K parameters and support training from scratch with limited training data. Here, we explore the scalability of PiDiNet with different model complexities as shown in Fig.~\ref{fig:scalability}. In order to compare with other approaches, the models are trained in two schemes, both use the BSDS500 training and validation set, while with or without mixing the VOC dataset during training. Metrics are recorded on BSDS500 test set. As expected, compared with the basic PiDiNet, smaller models suffer from lower network capacity and thus with degenerated performances in terms of both ODS and OIS scores. At the same time, training with more data constantly leads to higher accuracy. It is noted that the normal scale PiDiNet, can achieve the ODS and OIS scores at the same level as that recorded in the HED approach~\cite{xie2017holistically}, even when trained from scratch only using the BSDS500 dataset (\emph{i.e.}, 0.789 vs. 0.788 in ODS and 0.803 vs. 0.808 in OIS for PiDiNet vs. HED). However, with limited training data, widening the PiDiNet architecture may cause the overfitting problem, as shown in the declines in the second half of the curves. In the following experiments, we only use the tiny, small, and normal versions of PiDiNet, dubbed as PiDiNet-Tiny, PiDiNet-Small and PiDiNet respectively.

\begin{figure}[t!]
    \centering
    \includegraphics[width=1\linewidth]{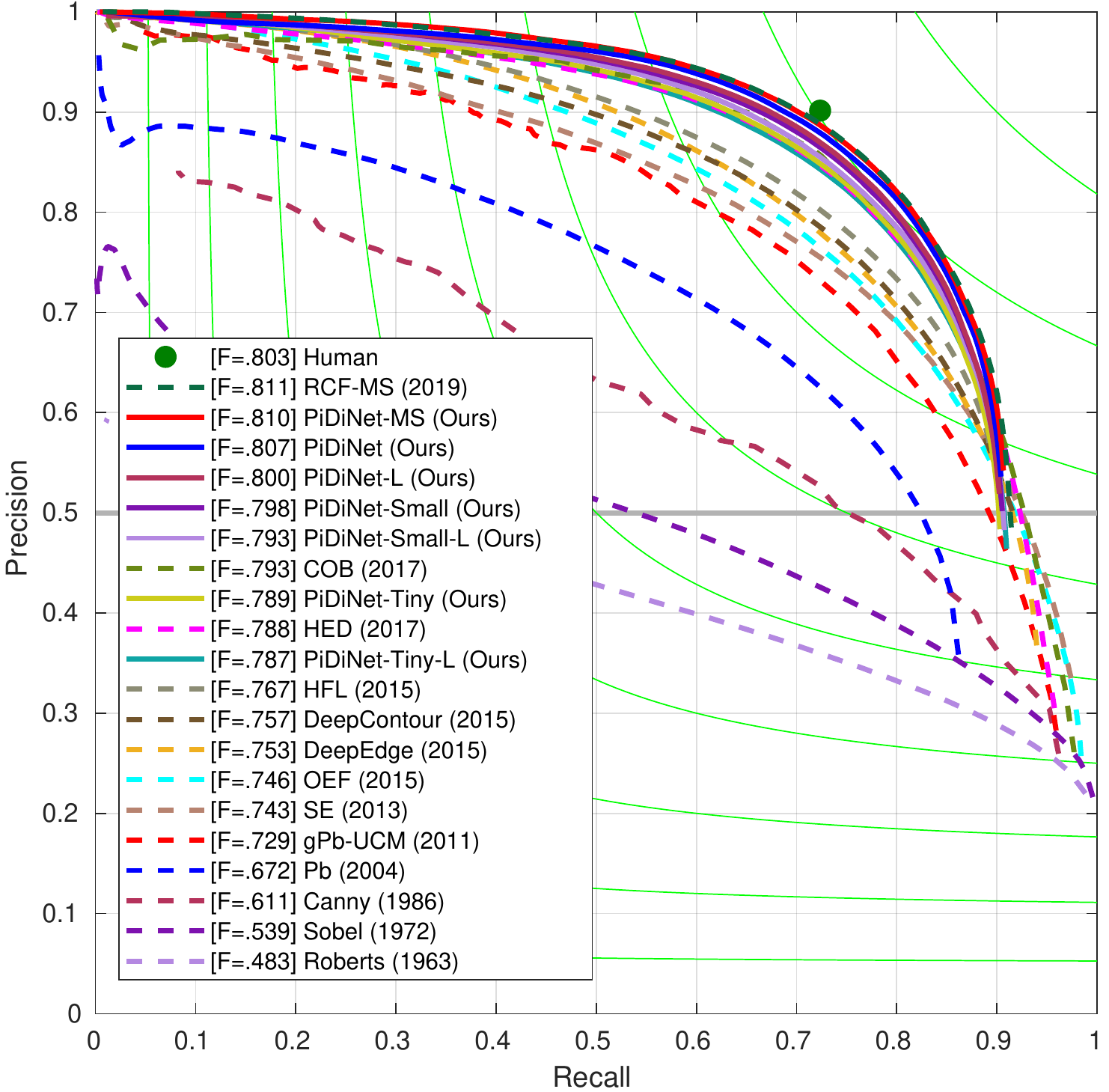}
    \caption{Precision-Recall curves of our models and some competitors on BSDS500 dataset.}
    \label{fig:bsds_pr}
\end{figure}

\subsection{Comparison with State-of-the-arts}

\begin{figure}[t!]
    \centering
    \includegraphics[width=1\linewidth]{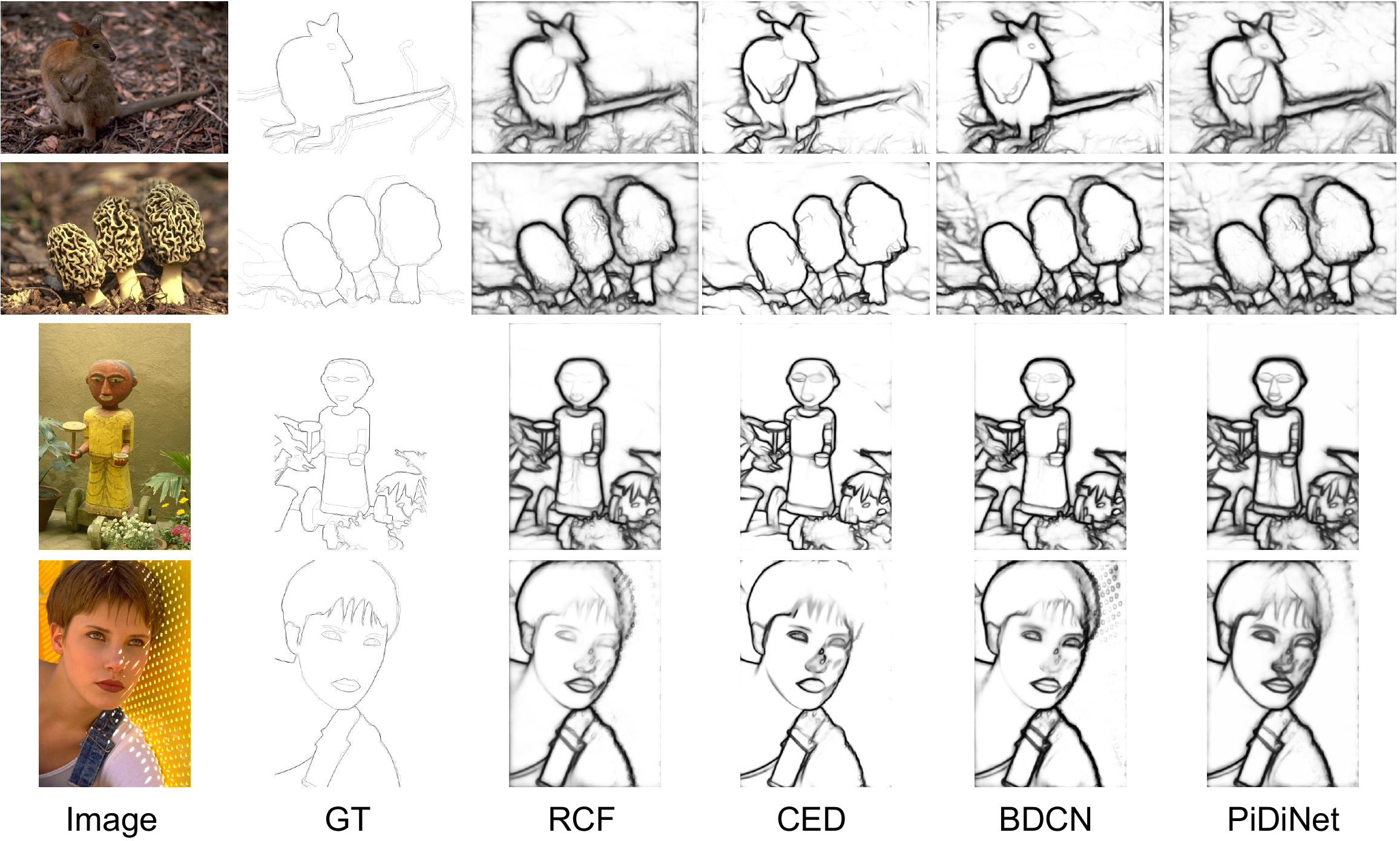}
    \caption{A qualitative comparison of network outputs with some other methods, including RCF~\cite{liu2019richer}, CED~\cite{wang2017ced} and BDCN~\cite{he2019bidirectional}.}
    \label{fig:bsds_visualization}
\end{figure}

\vspace{0.3em}
\noindent \textbf{On BSDS500 dataset.} \quad We compare our methods with prior edge detection approaches including both traditional ones and recently proposed CNN based ones, as summarized in Table~\ref{table:bsds} and Fig.~\ref{fig:bsds_pr}. Firstly, we notice that our baseline model can even achieve comparable results, \emph{i.e.}, with ODS of 0.798 and OIS of 0.816, already beating most CNN based models like CED~\cite{wang2017ced}, DeepBoundary~\cite{kokkinos2015deepboundary} and HED~\cite{xie2017holistically}. With PDC, PiDiNet can further boost the performance with ODS of 0.807, being the same level as the recently proposed RCF~\cite{liu2019richer} while still achieving nearly 100 FPS. The fastest version PiDiNet-Tiny-L, can also achieve comparable prediction performance with more than 200 FPS, further demonstrating the effectiveness of our methods. Noting all of our modes are trained from scratch using the same amount of training data as in RCF, LPCB, BDCN, \emph{etc}. (\emph{i.e.}, the training and validation set, mixed with the VOC dataset), without the ImageNet pretraining. We also show some qualitative results in Figure~\ref{fig:bsds_visualization}. A more detailed comparison in terms of network complexity, running efficiency and accuracy can be seen in Fig.~\ref{fig:efficient}. 

\vspace{0.3em}
\noindent \textbf{On NYUD dataset.} \quad The comparison results on the NYUD dataset are illustrated on Table~\ref{table:nyud}. Following the prior works, we get the `RGB-HHA' results by averaging the output edge maps from RGB image and HHA image to get the final edge map. The quantitative comparison shows that PiDiNets can still achieve highly comparable results among the state-of-the-art methods while being efficient. Please refer to the appendix for the Precision-Recall curves.

\vspace{0.3em}
\noindent \textbf{On Multicue dataset.} \quad We also record the evaluation results on Multicue dataset and the comparison results with other methods are shown on Table~\ref{table:multicue}. Still, PiDiNets achieve promising results with high efficiencies.


\begin{table}[t!]
\caption{Comparison with other methods on NYUD dataset. $^\ddagger$ indicates the speeds with our implementations based on a NVIDIA RTX 2080 Ti GPU.}
\vspace{-0.1in}
\begin{center}
\setlength{\tabcolsep}{0.025\linewidth}
\resizebox*{\linewidth}{!}{
\begin{tabular}{l|c|c|c|c|c|c|c}
\toprule[1pt]
Methods & ODS & OIS & ODS & OIS & ODS & OIS & FPS \\
\hline 
gPb-UCM~\cite{arbelaez2010bsds} & .632 & .661 & & & & & 1/360 \\
gPb+NG~\cite{gupta2013gpbng} & .687 & .716 & & & & & 1/375 \\
SE~\cite{dollar2014se} & .695 & .708 & & & & & 5 \\
SE+NG+~\cite{gupta2014seng} & .710 & .723 & & & & & 1/15 \\
\hline
\hline
& \multicolumn{2}{c|}{RGB} & \multicolumn{2}{c|}{HHA} & \multicolumn{2}{c|}{RGB-HHA} & \\
\hline
HED~\cite{xie2017holistically} & .720 & .734 & .682 & .695 & .746 & .761 & 62$^\ddagger$ \\
LPCB~\cite{deng2018lpcb} & .739 & .754 & .707 & .719 & .762 & .778 & - \\
RCF~\cite{liu2019richer} & .743 & .757 & .703 & .717 & .765 & .780 & 52$^\ddagger$ \\
AMH-Net~\cite{xu2018amhnet} & .744 & .758 & .716 & .729 & .771 & .786 & - \\
BDCN~\cite{he2019bidirectional} & .748 & .763 & .707 & .719 & .765 & .781 & 33$^\ddagger$ \\
\hline
PiDiNet & .733 & .747 & .715 & .728 & . 756 & .773 & 62$^\ddagger$ \\
PiDiNet-L & .728 & .741 & .709 & .722 & .754 & .770 & 88$^\ddagger$ \\
PiDiNet-Small & .726 & .741 & .705 & .719 & .750 & .767 & 115$^\ddagger$ \\
PiDiNet-Small-L & .721 & .736 & .701 & .713 & .746 & .763 & 165$^\ddagger$ \\
PiDiNet-Tiny & .721 & .736 & .700 & .714 & .745 & .763 & 140$^\ddagger$ \\
PiDiNet-Tiny-L & .714 & .729 & .693 & .706 & .741 & .759 & 206$^\ddagger$ \\
\bottomrule[1pt]
\end{tabular}
}
\end{center}
\label{table:nyud}
\vspace{-0.1in}
\end{table}


\begin{table}[t!]
\caption{Comparison with other methods on Multicue dataset. $^\ddagger$ indicates the speeds with our implementations based on a NVIDIA RTX 2080 Ti GPU.}
\vspace{0.05in}
\begin{center}
\setlength{\tabcolsep}{0.015\linewidth}
\resizebox*{\linewidth}{!}{
\begin{tabular}{l|c|c|c|c|c}
\toprule[1pt]
Method & \multicolumn{2}{c|}{Boundary} & \multicolumn{2}{c|}{Edge} & FPS \\
\hhline{~----~}
 & ODS & OIS & ODS & OIS & \\
\hline 
Human~\cite{mely2016multicue} & .760 (.017) & & .750 (.024) &  & \\
\hline
Multicue~\cite{mely2016multicue} & .720 (.014) & & .830 (.002) & & - \\
HED~\cite{xie2017holistically} & .814 (.011) & .822 (.008) & .851 (.014) & .864 (.011) & 18$^\ddagger$ \\
RCF~\cite{liu2019richer} & .817 (.004) & .825 (.005) & .857 (.004) & .862 (.004) & 15$^\ddagger$ \\
BDCN~\cite{he2019bidirectional} & .836 (.001) & .846 (.003) & .891 (.001) & .898 (.002) & 9$^\ddagger$ \\
\hline
PiDiNet & .818 (.003) & .830 (.005) & .855 (.007) & .860 (.005) & 17$^\ddagger$ \\
PiDiNet-L & .810 (.005) & .822 (.002) & .854 (.007) & .860 (.004) & 23$^\ddagger$ \\
PiDiNet-Small & .812 (.004) & .825 (.004) & .858 (.007) & .863 (.004) & 31$^\ddagger$ \\
PiDiNet-Small-L & .805 (.007) & .818 (.002) & .854 (.007) & .860 (.004) & 44$^\ddagger$ \\
PiDiNet-Tiny & .807 (.007) & .819 (.004) & .856 (.006) & .862 (.003) & 43$^\ddagger$ \\
PiDiNet-Tiny-L & .798 (.007) & .811 (.005) & .854 (.008) & .861 (.004) & 56$^\ddagger$ \\
\bottomrule[1pt]
\end{tabular}
}
\end{center}
\label{table:multicue}
\end{table}

\section{Conclusion}
In conclusion, the contribution in this paper is three-fold:
Firstly, we derive the pixel difference convolution which integrates the wisdom from the traditional edge detectors and the advantages of the deep CNNs, leading to robust and accurate edge detection.
Secondly, we propose a highly efficient architecture named PiDiNet based on pixel difference convolution, which are memory friendly and with high inference speed. Furthermore, PiDiNet can be trained from scratch only using limited data samples, while achieving human-level performances, breaking the convention that high performance CNN based edge detectors usually need a backbone pretrained on large scale dataset.
Thirdly, we conduct extensive experiments on BSDS500, NYUD, and Multicue datasets for edge detection. We believe that PiDiNet has created new state-of-the-art performances considering both accuracy and efficiency.

\vspace{0.5em}
\noindent \textbf{Future Work.} \quad 
As discussed in Section~\ref{sec:intro}, edge detection is a low level task for many mid- or high-level vision tasks like semantic segmentation and object detection. Also, some low level tasks like salient object detection may also benefit from the image boundary information. We hope pixel difference convolution and the proposed PiDiNet can go further and be useful in these related tasks. 

\vspace{0.5em}
\noindent \textbf{Acknowledgement.} \quad 
This work was partially supported by the Academy of Finland under grant 331883 and the National Natural Science Foundation of China under Grant 61872379, 62022091, and 71701205. The authors also wish to acknowledge CSC IT Center for Science, Finland, for computational resources.

{\small
\bibliographystyle{ieee_fullname}
\bibliography{main}

\begin{thebibliography}{10}\itemsep=-1pt

\bibitem{arbelaez2010bsds}
Pablo Arbelaez, Michael Maire, Charless Fowlkes, and Jitendra Malik.
\newblock Contour detection and hierarchical image segmentation.
\newblock {\em IEEE Transactions on Pattern Analysis and Machine Intelligence},
  33(5):898--916, 2011.

\bibitem{bertasius2015deepedge}
Gedas Bertasius, Jianbo Shi, and Lorenzo Torresani.
\newblock Deepedge: A multi-scale bifurcated deep network for top-down contour
  detection.
\newblock In {\em CVPR}, pages 4380--4389, 2015.

\bibitem{bertasius2015hfl}
Gedas Bertasius, Jianbo Shi, and Lorenzo Torresani.
\newblock High-for-low and low-for-high: Efficient boundary detection from deep
  object features and its applications to high-level vision.
\newblock In {\em ICCV}, pages 504--512, 2015.

\bibitem{bertasius2016semantic}
Gedas Bertasius, Jianbo Shi, and Lorenzo Torresani.
\newblock Semantic segmentation with boundary neural fields.
\newblock In {\em CVPR}, pages 3602--3610, 2016.

\bibitem{canny1986computational}
John Canny.
\newblock A computational approach to edge detection.
\newblock {\em IEEE Transactions on Pattern Analysis and Machine Intelligence},
  (6):679--698, 1986.

\bibitem{cheng2014bing}
Ming-Ming Cheng, Ziming Zhang, Wen-Yan Lin, and Philip Torr.
\newblock Bing: Binarized normed gradients for objectness estimation at 300fps.
\newblock In {\em CVPR}, pages 3286--3293, 2014.

\bibitem{deng2009imagenet}
Jia Deng, Wei Dong, Richard Socher, Li-Jia Li, Kai Li, and Li Fei-Fei.
\newblock Imagenet: A large-scale hierarchical image database.
\newblock In {\em CVPR}, pages 248--255. Ieee, 2009.

\bibitem{deng2018lpcb}
Ruoxi Deng, Chunhua Shen, Shengjun Liu, Huibing Wang, and Xinru Liu.
\newblock Learning to predict crisp boundaries.
\newblock In {\em ECCV}, pages 562--578, 2018.

\bibitem{dollar2014se}
Piotr Doll{\'a}r and C~Lawrence Zitnick.
\newblock Fast edge detection using structured forests.
\newblock {\em IEEE Transactions on Pattern Analysis and Machine Intelligence},
  37(8):1558--1570, 2015.

\bibitem{elder1998imageediting}
James~H Elder and Richard~M Goldberg.
\newblock Image editing in the contour domain.
\newblock In {\em Proceedings. 1998 IEEE Computer Society Conference on
  Computer Vision and Pattern Recognition (Cat. No. 98CB36231)}, pages
  374--381. IEEE, 1998.

\bibitem{ferrari2007groups}
Vittorio Ferrari, Loic Fevrier, Frederic Jurie, and Cordelia Schmid.
\newblock Groups of adjacent contour segments for object detection.
\newblock {\em IEEE Transactions on Pattern Analysis and Machine Intelligence},
  30(1):36--51, 2008.

\bibitem{fowlkes2002learningpb}
David R Martin Charless~C Fowlkes and Jitendra Malik.
\newblock Learning to detect natural image boundaries using brightness and
  texture.
\newblock {\em NeurIPS}, 2002.

\bibitem{gao2020100k}
Shang-Hua Gao, Yong-Qiang Tan, Ming-Ming Cheng, Chengze Lu, Yunpeng Chen, and
  Shuicheng Yan.
\newblock Highly efficient salient object detection with 100k parameters.
\newblock In {\em ECCV}, pages 702--721. Springer, 2020.

\bibitem{gupta2013gpbng}
Saurabh Gupta, Pablo Arbelaez, and Jitendra Malik.
\newblock Perceptual organization and recognition of indoor scenes from rgb-d
  images.
\newblock In {\em CVPR}, pages 564--571, 2013.

\bibitem{gupta2014seng}
Saurabh Gupta, Ross Girshick, Pablo Arbel{\'a}ez, and Jitendra Malik.
\newblock Learning rich features from rgb-d images for object detection and
  segmentation.
\newblock In {\em ECCV}, pages 345--360. Springer, 2014.

\bibitem{hallman2015oef}
Sam Hallman and Charless~C Fowlkes.
\newblock Oriented edge forests for boundary detection.
\newblock In {\em CVPR}, pages 1732--1740, 2015.

\bibitem{han2015deepcompression}
Song Han, Huizi Mao, and William~J. Dally.
\newblock Deep compression: Compressing deep neural network with pruning,
  trained quantization and huffman coding.
\newblock In {\em ICLR}, 2016.

\bibitem{he2019bidirectional}
Jianzhong He, Shiliang Zhang, Ming Yang, Yanhu Shan, and Tiejun Huang.
\newblock Bi-directional cascade network for perceptual edge detection.
\newblock In {\em CVPR}, pages 3828--3837, 2019.

\bibitem{he2016residual}
Kaiming He, Xiangyu Zhang, Shaoqing Ren, and Jian Sun.
\newblock Deep residual learning for image recognition.
\newblock In {\em CVPR}, pages 770--778, 2016.

\bibitem{howard2017mobilenets}
Andrew~G Howard, Menglong Zhu, Bo Chen, Dmitry Kalenichenko, Weijun Wang,
  Tobias Weyand, Marco Andreetto, and Hartwig Adam.
\newblock Mobilenets: Efficient convolutional neural networks for mobile vision
  applications.
\newblock {\em arXiv preprint arXiv:1704.04861}, 2017.

\bibitem{juefei2017lbc}
Felix Juefei-Xu, Vishnu Naresh~Boddeti, and Marios Savvides.
\newblock Local binary convolutional neural networks.
\newblock In {\em CVPR}, pages 19--28, 2017.

\bibitem{kingma2014adam}
Diederik~P. Kingma and Jimmy Ba.
\newblock Adam: {A} method for stochastic optimization.
\newblock In Yoshua Bengio and Yann LeCun, editors, {\em ICLR}, 2015.

\bibitem{kokkinos2015deepboundary}
Iasonas Kokkinos.
\newblock Pushing the boundaries of boundary detection using deep learning.
\newblock {\em arXiv preprint arXiv:1511.07386}, 2015.

\bibitem{kokkinos2017ubernet}
Iasonas Kokkinos.
\newblock Ubernet: Training a universal convolutional neural network for low-,
  mid-, and high-level vision using diverse datasets and limited memory.
\newblock In {\em CVPR}, pages 6129--6138, 2017.

\bibitem{li2019dabnet}
Gen Li and Joongkyu Kim.
\newblock Dabnet: Depth-wise asymmetric bottleneck for real-time semantic
  segmentation.
\newblock In {\em BMVC}, page 259. {BMVA} Press, 2019.

\bibitem{liu2020dynamicintegration}
Jiang-Jiang Liu, Qibin Hou, and Ming-Ming Cheng.
\newblock Dynamic feature integration for simultaneous detection of salient
  object, edge, and skeleton.
\newblock {\em IEEE Transactions on Image Processing}, 29:8652--8667, 2020.

\bibitem{liu2019bow}
Li Liu, Jie Chen, Paul Fieguth, Guoying Zhao, Rama Chellappa, and Matti
  Pietik{\"a}inen.
\newblock From bow to cnn: Two decades of texture representation for texture
  classification.
\newblock {\em International Journal of Computer Vision}, 127(1):74--109, 2019.

\bibitem{liu2011sorted}
Li Liu, Paul Fieguth, Gangyao Kuang, and Hongbin Zha.
\newblock Sorted random projections for robust texture classification.
\newblock In {\em ICCV}, pages 391--398. IEEE, 2011.

\bibitem{liu2020deep}
Li Liu, Wanli Ouyang, Xiaogang Wang, Paul Fieguth, Jie Chen, Xinwang Liu, and
  Matti Pietik{\"a}inen.
\newblock Deep learning for generic object detection: A survey.
\newblock {\em International Journal of Computer Vision}, 128(2):261--318,
  2020.

\bibitem{liu2012extended}
Li Liu, Lingjun Zhao, Yunli Long, Gangyao Kuang, and Paul Fieguth.
\newblock Extended local binary patterns for texture classification.
\newblock {\em Image and Vision Computing}, 30(2):86--99, 2012.

\bibitem{liu2019richer}
Yun Liu, Ming-Ming Cheng, Xiaowei Hu, Jia-Wang Bian, Le Zhang, Xiang Bai, and
  Jinhui Tang.
\newblock Richer convolutional features for edge detection.
\newblock {\em IEEE Transactions on Pattern Analysis and Machine Intelligence},
  41(8):1939--1946, 2019.

\bibitem{liu2016relaxed}
Yu Liu and Michael~S Lew.
\newblock Learning relaxed deep supervision for better edge detection.
\newblock In {\em CVPR}, pages 231--240, 2016.

\bibitem{long2015fully}
Jonathan Long, Evan Shelhamer, and Trevor Darrell.
\newblock Fully convolutional networks for semantic segmentation.
\newblock In {\em CVPR}, pages 3431--3440, 2015.

\bibitem{luan2018gabor}
Shangzhen Luan, Chen Chen, Baochang Zhang, Jungong Han, and Jianzhuang Liu.
\newblock Gabor convolutional networks.
\newblock {\em IEEE Transactions on Image Processing}, 27(9):4357--4366, 2018.

\bibitem{ma2018shufflenetv2}
Ningning Ma, Xiangyu Zhang, Hai-Tao Zheng, and Jian Sun.
\newblock Shufflenet v2: Practical guidelines for efficient cnn architecture
  design.
\newblock In {\em ECCV}, pages 116--131, 2018.

\bibitem{maninis2016cob}
Kevis-Kokitsi Maninis, Jordi Pont-Tuset, Pablo Arbel{\'a}ez, and Luc Van~Gool.
\newblock Convolutional oriented boundaries.
\newblock In {\em ECCV}, pages 580--596. Springer, 2016.

\bibitem{martin2004pb}
David~R Martin, Charless~C Fowlkes, and Jitendra Malik.
\newblock Learning to detect natural image boundaries using local brightness,
  color, and texture cues.
\newblock {\em IEEE Transactions on Pattern Analysis and Machine Intelligence},
  26(5):530--549, 2004.

\bibitem{mehta2019espnetv2}
Sachin Mehta, Mohammad Rastegari, Linda Shapiro, and Hannaneh Hajishirzi.
\newblock Espnetv2: A light-weight, power efficient, and general purpose
  convolutional neural network.
\newblock In {\em CVPR}, pages 9190--9200, 2019.

\bibitem{mely2016multicue}
David~A M{\'e}ly, Junkyung Kim, Mason McGill, Yuliang Guo, and Thomas Serre.
\newblock A systematic comparison between visual cues for boundary detection.
\newblock {\em Vision Research}, 120:93--107, 2016.

\bibitem{mottaghi2014voc}
Roozbeh Mottaghi, Xianjie Chen, Xiaobai Liu, Nam-Gyu Cho, Seong-Whan Lee, Sanja
  Fidler, Raquel Urtasun, and Alan Yuille.
\newblock The role of context for object detection and semantic segmentation in
  the wild.
\newblock In {\em CVPR}, pages 891--898, 2014.

\bibitem{muthukrishnan2011edgeimageseg}
R Muthukrishnan and Miyilsamy Radha.
\newblock Edge detection techniques for image segmentation.
\newblock {\em International Journal of Computer Science \& Information
  Technology}, 3(6):259, 2011.

\bibitem{ojala2002lbp}
Timo Ojala, Matti Pietikainen, and Topi Maenpaa.
\newblock Multiresolution gray-scale and rotation invariant texture
  classification with local binary patterns.
\newblock {\em IEEE Transactions on Pattern Analysis and Machine Intelligence},
  24(7):971--987, 2002.

\bibitem{paszke2016enet}
Adam Paszke, Abhishek Chaurasia, Sangpil Kim, and Eugenio Culurciello.
\newblock Enet: A deep neural network architecture for real-time semantic
  segmentation.
\newblock {\em arXiv preprint arXiv:1606.02147}, 2016.

\bibitem{paszke2019pytorch}
Adam Paszke, Sam Gross, Francisco Massa, Adam Lerer, James Bradbury, Gregory
  Chanan, Trevor Killeen, Zeming Lin, Natalia Gimelshein, Luca Antiga, Alban
  Desmaison, Andreas K{\"{o}}pf, Edward Yang, Zachary DeVito, Martin Raison,
  Alykhan Tejani, Sasank Chilamkurthy, Benoit Steiner, Lu Fang, Junjie Bai, and
  Soumith Chintala.
\newblock Pytorch: An imperative style, high-performance deep learning library.
\newblock In {\em NeurIPS}, pages 8024--8035, 2019.

\bibitem{poma2020dense}
Xavier~Soria Poma, Edgar Riba, and Angel Sappa.
\newblock Dense extreme inception network: Towards a robust cnn model for edge
  detection.
\newblock In {\em WACV}, pages 1923--1932, 2020.

\bibitem{prewitt1970object}
Judith~MS Prewitt.
\newblock Object enhancement and extraction.
\newblock {\em Picture processing and Psychopictorics}, 10(1):15--19, 1970.

\bibitem{shen2015deepcontour}
Wei Shen, Xinggang Wang, Yan Wang, Xiang Bai, and Zhijiang Zhang.
\newblock Deepcontour: A deep convolutional feature learned by positive-sharing
  loss for contour detection.
\newblock In {\em CVPR}, pages 3982--3991, 2015.

\bibitem{shi2000nyud}
Jianbo Shi and Jitendra Malik.
\newblock Normalized cuts and image segmentation.
\newblock {\em IEEE Transactions on Pattern Analysis and Machine Intelligence},
  22(8):888--905, 2000.

\bibitem{simonyan2014very}
Karen Simonyan and Andrew Zisserman.
\newblock Very deep convolutional networks for large-scale image recognition.
\newblock In {\em ICLR}, 2015.

\bibitem{sobel19683x3}
Irwin Sobel and Gary Feldman.
\newblock A 3x3 isotropic gradient operator for image processing.
\newblock {\em A Talk at The Stanford Artificial Project in}, pages 271--272,
  1968.

\bibitem{su2020dynamic}
Zhuo Su, Linpu Fang, Wenxiong Kang, Dewen Hu, Matti Pietik{\"a}inen, and Li
  Liu.
\newblock Dynamic group convolution for accelerating convolutional neural
  networks.
\newblock In {\em ECCV}, pages 138--155. Springer, 2020.

\bibitem{su2019bird}
Zhuo Su, Matti Pietik{\"a}inen, and Li Liu.
\newblock Bird: Learning binary and illumination robust descriptor for face
  recognition.
\newblock In {\em BMVC}, page 102, 2019.

\bibitem{torre1986edge}
Vincent Torre and Tomaso~A Poggio.
\newblock On edge detection.
\newblock {\em IEEE Transactions on Pattern Analysis and Machine Intelligence},
  (2):147--163, 1986.

\bibitem{uijlings2013selective}
Jasper~RR Uijlings, Koen~EA Van De~Sande, Theo Gevers, and Arnold~WM Smeulders.
\newblock Selective search for object recognition.
\newblock {\em International Journal of Computer Vision}, 104(2):154--171,
  2013.

\bibitem{wang2017ced}
Yupei Wang, Xin Zhao, and Kaiqi Huang.
\newblock Deep crisp boundaries.
\newblock In {\em CVPR}, pages 3892--3900, 2017.

\bibitem{wibisono2020fined}
Jan~Kristanto Wibisono and Hsueh-Ming Hang.
\newblock Fined: Fast inference network for edge detection.
\newblock {\em arXiv preprint arXiv:2012.08392}, 2020.

\bibitem{wibisono2020traditional}
Jan~Kristanto Wibisono and Hsueh-Ming Hang.
\newblock Traditional method inspired deep neural network for edge detection.
\newblock In {\em ICIP}, pages 678--682. IEEE, 2020.

\bibitem{wu2020cgnet}
Tianyi Wu, Sheng Tang, Rui Zhang, Juan Cao, and Yongdong Zhang.
\newblock Cgnet: A light-weight context guided network for semantic
  segmentation.
\newblock {\em IEEE Transactions on Image Processing}, 30:1169--1179, 2020.

\bibitem{xiaofeng2012scg}
Ren Xiaofeng and Liefeng Bo.
\newblock Discriminatively trained sparse code gradients for contour detection.
\newblock In {\em NeurIPS}, pages 584--592, 2012.

\bibitem{xie2017holistically}
Saining Xie and Zhuowen Tu.
\newblock Holistically-nested edge detection.
\newblock {\em International Journal of Computer Vision}, 125(1-3):3--18, 2017.

\bibitem{xu2018amhnet}
Dan Xu, Wanli Ouyang, Xavier Alameda-Pineda, Elisa Ricci, Xiaogang Wang, and
  Nicu Sebe.
\newblock Learning deep structured multi-scale features using attention-gated
  crfs for contour prediction.
\newblock In {\em NeurIPS}, pages 3961--3970, 2017.

\bibitem{yang2016cedn}
Jimei Yang, Brian Price, Scott Cohen, Honglak Lee, and Ming-Hsuan Yang.
\newblock Object contour detection with a fully convolutional encoder-decoder
  network.
\newblock In {\em CVPR}, pages 193--202, 2016.

\bibitem{yu2018bisenet}
Changqian Yu, Jingbo Wang, Chao Peng, Changxin Gao, Gang Yu, and Nong Sang.
\newblock Bisenet: Bilateral segmentation network for real-time semantic
  segmentation.
\newblock In {\em ECCV}, pages 325--341, 2018.

\bibitem{yu2021dual}
Zitong Yu, Yunxiao Qin, Hengshuang Zhao, Xiaobai Li, and Guoying Zhao.
\newblock Dual-cross central difference network for face anti-spoofing.
\newblock In {\em IJCAI}, 2021.

\bibitem{yu2020fas}
Zitong Yu, Jun Wan, Yunxiao Qin, Xiaobai Li, Stan~Z Li, and Guoying Zhao.
\newblock Nas-fas: Static-dynamic central difference network search for face
  anti-spoofing.
\newblock {\em IEEE Transactions on Pattern Analysis and Machine Intelligence},
  2021.

\bibitem{yu2020cdc}
Zitong Yu, Chenxu Zhao, Zezheng Wang, Yunxiao Qin, Zhuo Su, Xiaobai Li, Feng
  Zhou, and Guoying Zhao.
\newblock Searching central difference convolutional networks for face
  anti-spoofing.
\newblock In {\em CVPR}, pages 5295--5305, 2020.

\bibitem{yu2021searching}
Zitong Yu, Benjia Zhou, Jun Wan, Pichao Wang, Haoyu Chen, Xin Liu, Stan~Z Li,
  and Guoying Zhao.
\newblock Searching multi-rate and multi-modal temporal enhanced networks for
  gesture recognition.
\newblock {\em IEEE Transactions on Image Processing}, 2021.

\end{thebibliography}
}

\clearpage

\section{Appendix}

\subsection{Converting Pixel Difference Convolution (PDC) to Vanilla Convolution}

The main goal of the conversion is to make PDC as fast and memory efficient as as the vanilla convolution. 
As introduced in the main paper, the formulations of vanilla convolution and PDC can be written as:

\begin{align}
    y &= f(\pmb{x}, \pmb{\theta}) = \sum_{i=1}^{k\times k}w_{i}\cdot x_{i}, \;\;\;\;\;\;\; \text{(vanilla convolution)}\label{eq:vanilla} \\
    y &= f(\triangledown\pmb{x}, \pmb{\theta}) = \sum_{(x_i, x_i')\in \pmb{\mathcal{P}}}w_{i}\cdot (x_i - x_i'), \;\;\;\;\;\;\, \text{(PDC)} \label{eq: pdc}
\end{align}
where, $x_i$ and $x_i'$ are the pixels in the current input local patch, $w_i$ is the weight in the $k \times k$ convolution kernel. $\pmb{\mathcal{P}} = \{(x_1, x_1'), (x_2, x_2'), ..., (x_m, x_m')\}$ is the set of pixel pairs picked from the local patch, and $m\le k\times k$. 

The conversion from PDC to vanilla convolution can be done in both the training and inference phases.

\vspace{0.3em}
\noindent  \textbf{Conversion in the Training Phase.} \quad Eq.~\ref{eq: pdc} can be transformed to fit the form of Eq.~\ref{eq:vanilla}, according to the selection strategies of the pixel pairs. Correspondingly, PDC can be converted to vanilla convolution by firstly transforming the kernel weights to a new set of kernel weights, followed by a vanilla convolutional operation. We will discuss Central PDC (CPDC), Angular PDC (APDC) and Radial PDC (RPDC) respectively.
The selection strategies of pixel pairs in the three PDC instances are shown in Fig.~\ref{fig:cpdc}, Fig.~\ref{fig:apdc} and Fig.~\ref{fig:rpdc}. The transformations of the equations are as follows.

\vspace{0.3em}
\noindent For CPDC (Fig.~\ref{fig:cpdc}):

\begin{figure}[t!]
    \centering
    \includegraphics[width=0.95\linewidth]{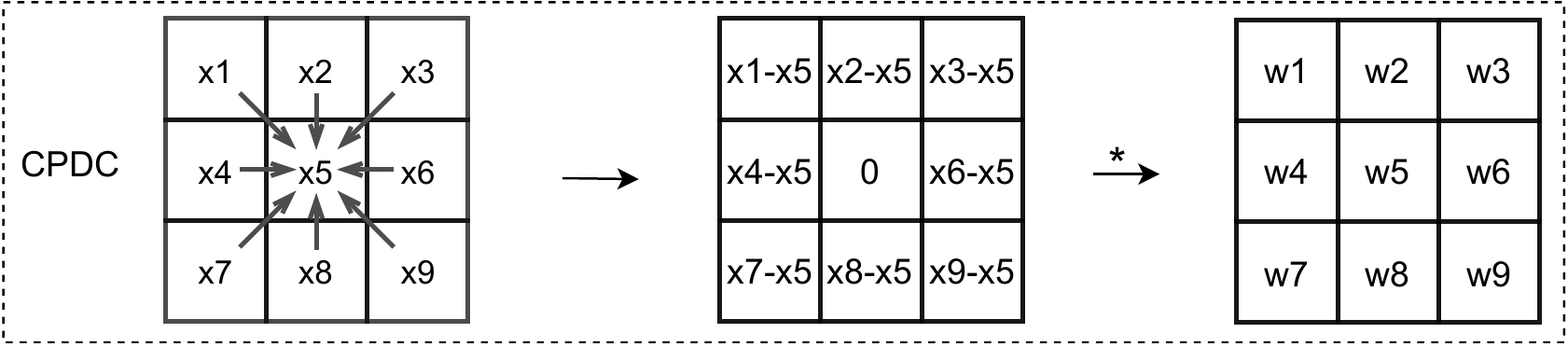}
    \caption{Selection of pixel pairs and convolution in CPDC.}
    \label{fig:cpdc}
\end{figure}

{\small 
\begin{align}
    y =& w_{1}\cdot (x_1 - x_5) + w_2\cdot (x_2 - x_5)+w_3\cdot (x_3 - x_5)\nonumber\\
    & + w_4\cdot (x_4-x_5) + w_6\cdot (x_6 - x_5) + w_7\cdot (x_7 - x_5)\nonumber \\
    & + w_8\cdot (x_8 - x_5) + w_9\cdot (x_9 - x_5)\nonumber \\
    =&w_1\cdot x_1 + w_2\cdot x_2 + w_3\cdot x_3 +\nonumber \\
    & + w_4\cdot x_4 + w_6\cdot x_6 + w_7\cdot x_7 +\nonumber \\
    & + w_8\cdot x_8 + w_9\cdot x_9\nonumber \\
    & + (-\sum_{i=\{1,2,3,4,6,7,8,9\}}w_i)\cdot x5\nonumber \\
    =&\hat{w}_1\cdot x_1 + \hat{w}_2\cdot x_2 + \hat{w}_3\cdot x_3 + ... =\sum \hat{w}_i\cdot x_i
\end{align}
}

\begin{figure}[t!]
    \centering
    \includegraphics[width=0.95\linewidth]{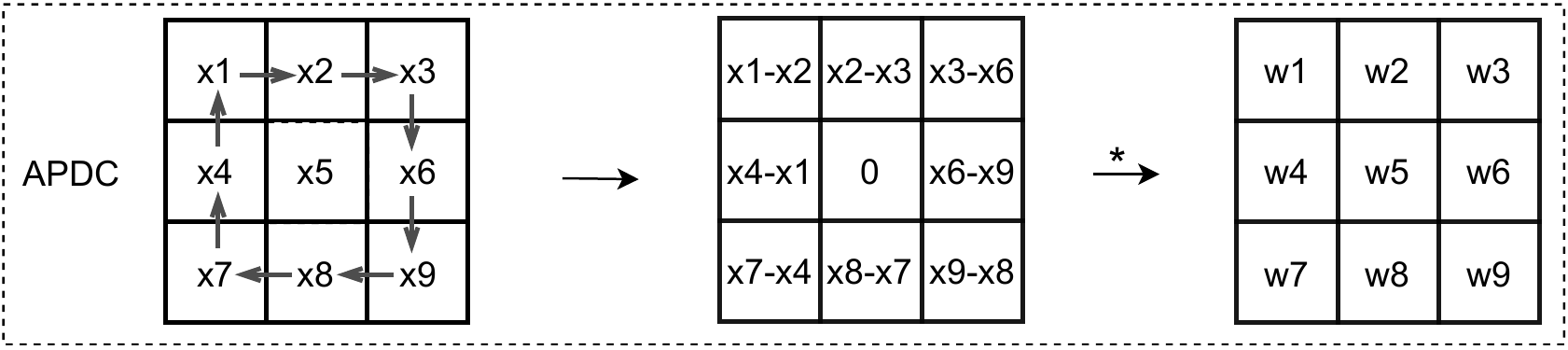}
    \caption{Selection of pixel pairs and convolution in APDC.}
    \label{fig:apdc}
\end{figure}

\vspace{0.3em}
\noindent For APDC (Fig.~\ref{fig:apdc}):

{\small
\begin{align}
    y =& w_{1}\cdot (x_1 - x_2) + w_2\cdot (x_2 - x_3)+w_3\cdot (x_3 - x_6)\nonumber\\
    & + w_4\cdot (x_4-x_1) + w_6\cdot (x_6 - x_9) + w_7\cdot (x_7 - x_4)\nonumber \\
    & + w_8\cdot (x_8 - x_7) + w_9\cdot (x_9 - x_8)\nonumber \\
    =& (w_1 - w_4)\cdot x_1 + (w_2 - w_1)\cdot x_2 + (w_3-w_2)\cdot x_3 \nonumber \\
    & + (w_4 - w_7)\cdot x_4 + (w_6 - w_3)\cdot x_6 + (w_7 - w_8)\cdot x_7 \nonumber \\
    & + (w_8 - w_9)\cdot x_8 + (w_9 - x_6)\cdot x_9\nonumber \\
    & + 0\cdot x_5\nonumber \\
    =& \hat{w}_1\cdot x_1 + \hat{w}_2\cdot x_2 + \hat{w}_3\cdot x_3 + ... =\sum \hat{w}_i\cdot x_i
\end{align}
}

\begin{figure}[t!]
    \centering
    \includegraphics[width=0.95\linewidth]{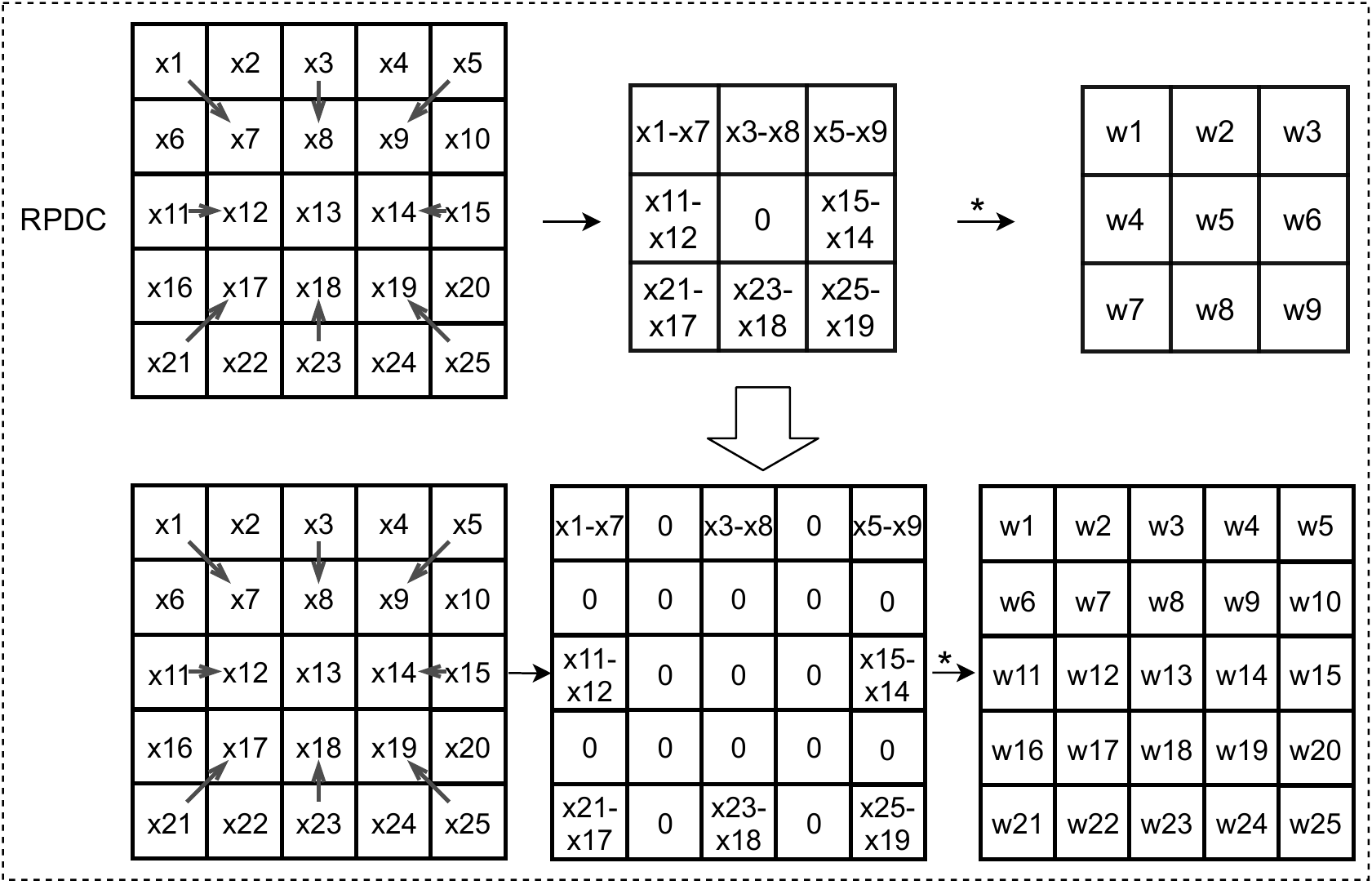}
    \caption{Selection of pixel pairs and convolution in RPDC.}
    \label{fig:rpdc}
\end{figure}

\vspace{0.3em}
\noindent For RPDC (Fig.~\ref{fig:rpdc}):

{\small
\begin{align}
    y =& w_{1}\cdot (x_1 - x_7) + w_3\cdot (x_3 - x_8)+w_5\cdot (x_5 - x_9)\nonumber\\
    & + w_{11}\cdot (x_{11}-x_{12}) + w_{15}\cdot (x_{15} - x_{14})\nonumber \\
    &  + w_{21}\cdot (x_{21} - x_{17}) + w_{23}\cdot (x_{23} - x_{18})\nonumber \\
    &  + w_{25}\cdot (x_{25} - x_{19})\nonumber\\
    =&w_1\cdot x_1 + w_3\cdot x_3 + w_5\cdot x_5\nonumber \\
    & + (-w_1)\cdot x_7 + (-w_3)\cdot x_8 + (-w_5)\cdot x_9 +\nonumber \\
    & + w_{11}\cdot x_{11} + (-w_{11})\cdot x_{12} + (-w_{15})\cdot x_{14}\nonumber\\
    & + w_{15}\cdot x_{15} + (-w_{21})\cdot x_{17} + (-w_{23})\cdot x_{18}\nonumber\\
    & + (-w_{25})\cdot x_{19} + w_{21}\cdot x_{21} + w_{23}\cdot x_{23}\nonumber\\
    & + w_{25}\cdot x_{25} + \sum_{i=\{2,4,6,10,13,16,20,22,24\}}0\cdot x_i\nonumber\\
    =&\hat{w}_1\cdot x_1 + \hat{w}_2\cdot x_2 + \hat{w}_3\cdot x_3 + ... =\sum \hat{w}_i\cdot x_i
\end{align}
}

The RPDC is converted to a vanilla convolution with kernel size $5\times 5$.

\vspace{0.3em}
\noindent  \textbf{Conversion in the Inference Phase.} \quad After training, instead of saving the original weights $w_i$, we directly save the new set of weights $\hat{w}_i$. Therefore, during inference, all the convolutional operations are vanilla convolutions.

\begin{figure}[t!]
    \centering
    \includegraphics[width=0.9\linewidth]{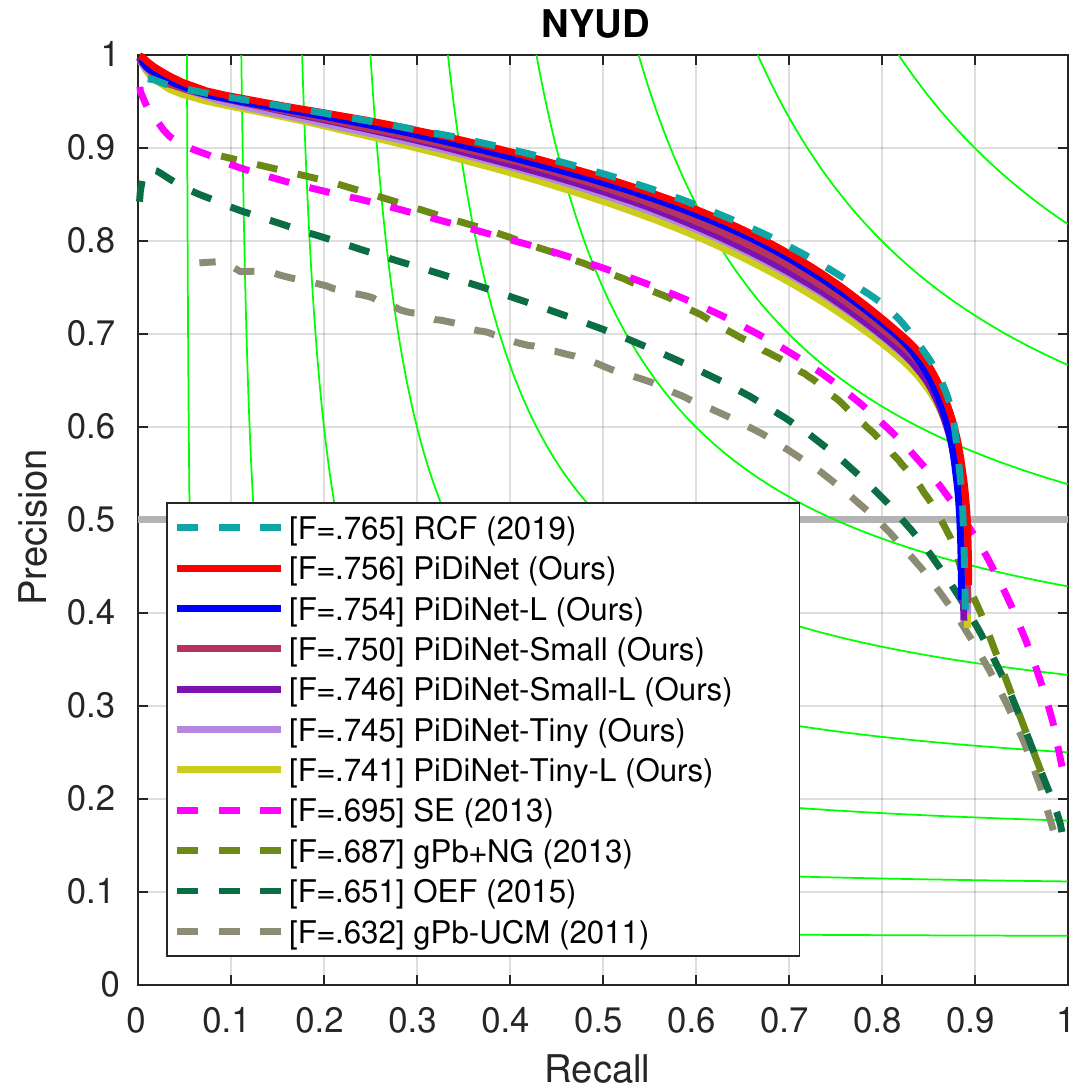}
    \caption{Precision-Recall curves of our models and some competitors on NYUD dataset.}
    \label{fig:nyud_pr}
\end{figure}

\subsection{Precision-Recall Curves on NYUD Dataset}

The Precision-Reall curves of our methods and other approaches on NYUD dataset~\cite{shi2000nyud} are shown in Fig.~\ref{fig:nyud_pr}. The compared methods include RCF~\cite{liu2019richer}, SE~\cite{dollar2014se}, gPb+NG~\cite{gupta2013gpbng}, gPb-UCM~\cite{arbelaez2010bsds} and OEF~\cite{hallman2015oef}.

\begin{figure*}[t!]
    \centering
    \includegraphics[width=0.97\linewidth]{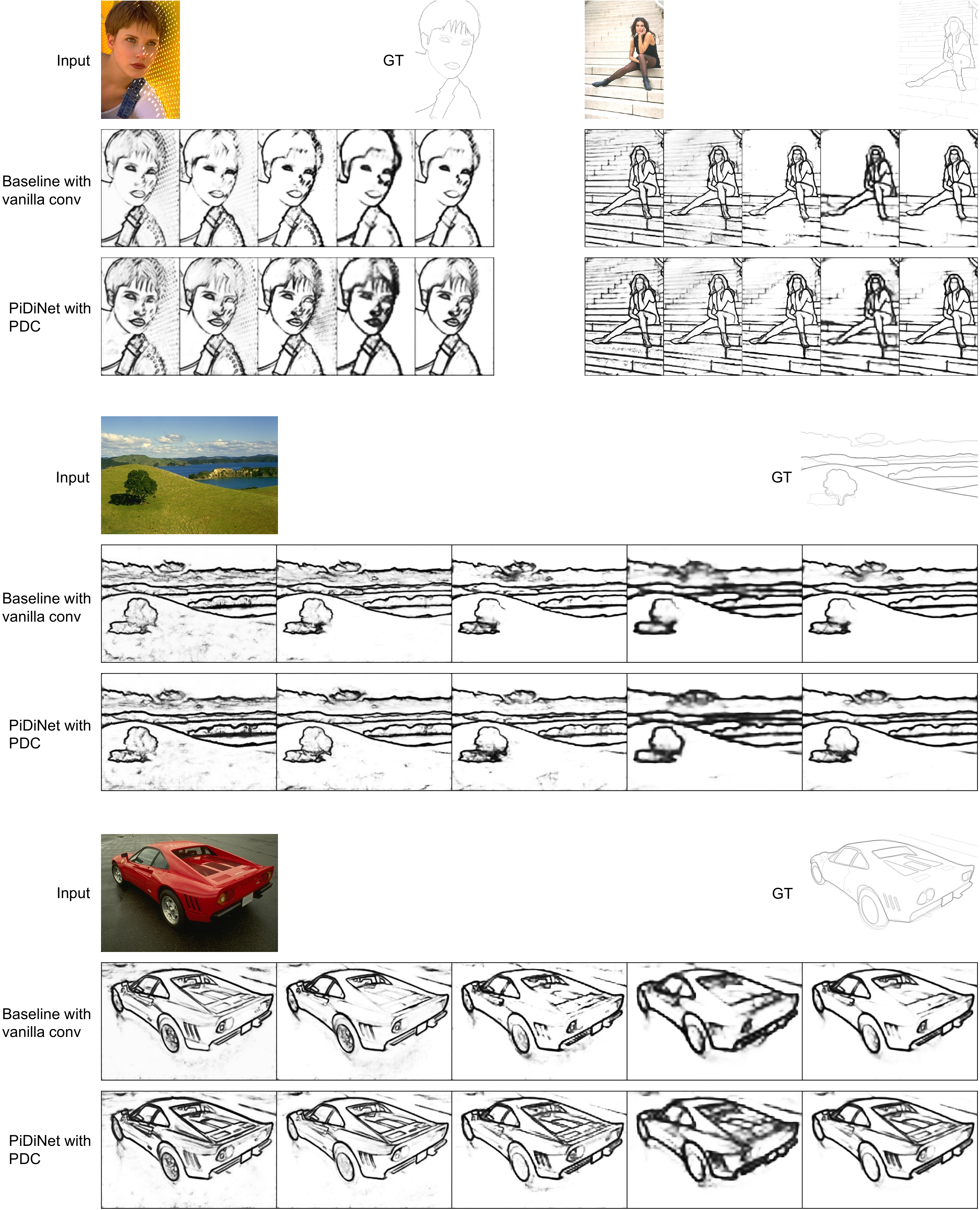}
    \caption{For each case, Top: input and ground truth image; Middle: edge maps from stage 1, 2, 3, 4 respectively and the final edge map, generated from the baseline architecture, Bottom: Corresponding edge maps generated from PiDiNet. Both the baseline architecture and PiDiNet were trained only using the BSDS500 dataset~\cite{arbelaez2010bsds}. Compared with the baseline, we can see that PiDiNet can detect more useful boundaries (\emph{e.g.}, bangs, stairs, the contour of the tree, the characteristic textures of the car).}
    \label{fig:stages}
\end{figure*}

\begin{figure*}[t!]
    \centering
    \includegraphics[width=1\linewidth]{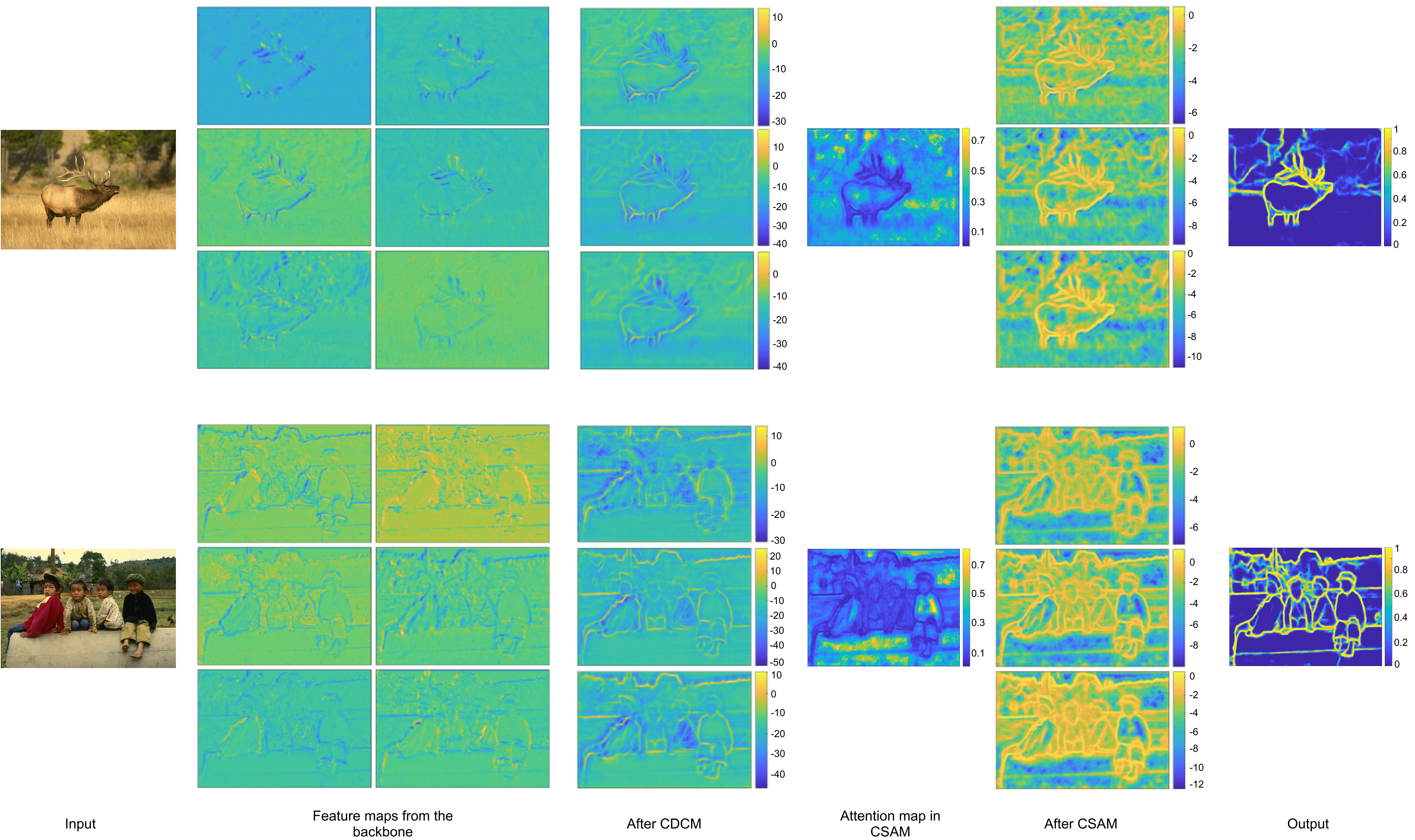}
    \caption{CDCM and CSAM can further refine the feature maps with multi-scale feature extraction and the sample adaptive spatial attention mechanism. Note that in the attention maps generated by CSAM, pixels in the background show higher intensities. This makes sense as the background pixels after CDCM have negative values, hence they will be additionally suppressed through CSAM.}
    \label{fig:maps}
\end{figure*}

\subsection{Visualization}

\vspace{0.3em}
\noindent  \textbf{Edge Maps.} \quad The edge maps generated from the baseline architecture and PiDiNet are shown in Fig.~\ref{fig:stages}. Both models were trained using only the BSDS500 dataset without the mixed VOC dataset~\cite{mottaghi2014voc}. From the figure, it is proved that PDC can help PiDiNet effectively capture more useful boundaries, with the ability to extract rich gradient information that facilitates edge detection. 

\vspace{0.3em}
\noindent  \textbf{Intermediate Feature Maps.} \quad We also visualize the intermediate feature maps extracted from PiDiNet, to qualitatively demonstrate the effectiveness of the compact dilation convolution based module (CDCM) and the compact spatial attention module (CSAM), which are shown in Fig.~\ref{fig:maps}. It is concluded that both CDCM and CSAM take a positive role in PiDiNet on the edge detection task.

\end{document}